\pdfoutput=1
\documentclass[11pt]{article}

\usepackage[final]{acl}

\usepackage{times}
\usepackage{latexsym}

\usepackage[T1]{fontenc}
\usepackage[utf8]{inputenc}
\usepackage{graphicx}

\usepackage{microtype}

\usepackage{inconsolata}

\usepackage{graphicx}

\usepackage{blindtext}
\newcommand\blfootnote[1]{%
\begingroup
\renewcommand\thefootnote{}\footnote{#1}%
\addtocounter{footnote}{-1}%
\endgroup
}

\usepackage{adjustbox}
\usepackage{subfig}
\usepackage{soul}
\usepackage{colortbl}
\usepackage{geometry}
\usepackage[breakable]{tcolorbox}
\usepackage{tabularx}
\usepackage{longtable}
\usepackage{array}
\usepackage{booktabs}
\usepackage{amsmath}
\usepackage{float}
\usepackage{wrapfig}
\usepackage[htt]{hyphenat}
\title{Chatbot To Help Patients Understand Their Health}

\author{
Won Seok Jang\thanks{Equal contribution}~$^{1,2}$, 
Hieu Tran\footnotemark[1]~$^{1,3}$, 
Manav Mistry~$^{1}$,
SaiKiran Gandluri~$^{1}$, 
\textbf{Yifan Zhang}~$^{1,2}$\\
\textbf{Sharmin Sultana}~$^{1,2}$,
\textbf{Sunjae Kwon}~$^{1,3}$,
\textbf{Yuan Zhang}~$^{2}$,
\textbf{Zonghai Yao}\thanks{Co-corresponding authors}~$^{1,3}$,
\textbf{Hong Yu}\footnotemark[2]~$^{1,2,3}$ \\
$^{1}$Center for Healthcare Organization and Implementation Research, VA Bedford Health Care  \\
$^{2}$Miner School of Computer and Information Sciences, University of Massachusetts Lowell \\
$^{3}$Manning College of Information and Computer Sciences, University of Massachusetts Amherst \\
\texttt{\{WonSeok\_Jang, Yifan\_Zhang, Sharmin\_Sultana,Yuan\_Zhang,Hong\_Yu\}@uml.edu} \\
\texttt{\{Manav\_Mistry,Saikiran\_gandluri\}@student.uml.edu} \\
\texttt{\{hieutran,sunjaekwon,zonghaiyao\}@umass.edu}
}

\begin{document}
\maketitle

\begin{abstract}
Patients must possess the knowledge necessary to actively participate in their care. 
We present NoteAid-Chatbot, a conversational AI that promotes patient understanding via a novel ‘learning as conversation’ framework, 
built on a multi-agent large language model (LLM) and reinforcement learning (RL) setup without human-labeled data.
NoteAid-Chatbot was built on a lightweight 3B-parameter LLaMA 3.2 model  trained in two stages: initial supervised fine-tuning on conversational data synthetically generated using medical conversation strategies, followed by RL with rewards derived from patient understanding assessments in simulated hospital discharge scenarios. 
Our evaluation, which includes comprehensive human-aligned assessments and case studies, demonstrates that NoteAid-Chatbot exhibits key emergent behaviors critical for patient education—such as clarity, relevance, and structured dialogue—even though it received no explicit supervision for these attributes. Our results show that even simple Proximal Policy Optimization (PPO)-based reward modeling can successfully train lightweight, domain-specific chatbots to handle multi-turn interactions, incorporate diverse educational strategies, and meet nuanced communication objectives. Our Turing test demonstrates that NoteAid-Chatbot surpasses non-expert human. Although our current focus is on healthcare, the framework we present illustrates the feasibility and promise of applying low-cost, PPO-based RL to realistic, open-ended conversational domains—broadening the applicability of RL-based alignment methods.

\end{abstract}

\section{Introduction}
\begin{figure*}[ht]
    \centering
    % \includegraphics[width=0.8\linewidth, trim= 0cm 0cm cm 0cm, clip]{stage1-2.pdf}
    % \caption{Data preparation and supervised fine-tuning stage}
    % \label{fig:sft-stage}
% \end{figure}
% \begin{figure}
    % \vspace{0}
    \centering
    \includegraphics[width=\linewidth, trim=0cm 0cm 0cm 0cm,clip]{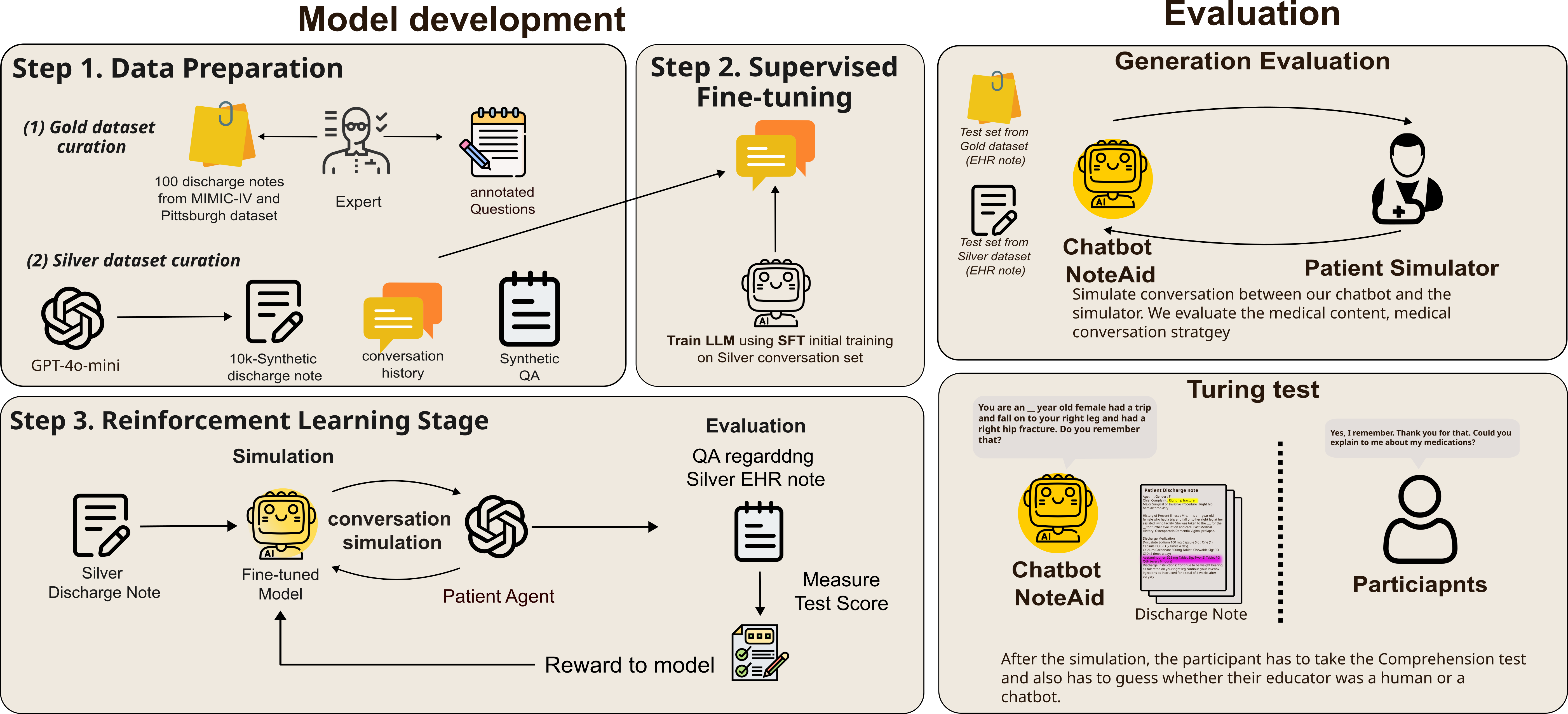}
    % \caption{The NoteAid Chatbot pipeline. First starts with (i) data preparation and (ii) supervised fine-tuning. Finally for further alignment, (iii) reinforcement learning.}
    \caption{Overview of our multi-agent framework and interactive patient education experiment.
    \textbf{(Left: Model development)} The NoteAid-Chatbot training pipeline. We first construct a two datasets: 1) Gold dataset that consists real-world EHR notes and questionnaires annotated by experts, 2) Silver dataset which is synthetic dataset (EHR notes, Conversation records, questionnaires) generated using six medical content criteria and medical conversation strategy. We apply supervised fine-tuning on this conversation dataset to build a baseline chatbot model. Leveraging the Silver dataset, we align the chatbot via reinforcement learning (PPO), where the Chatbot NoteAid interacts with the patient agent (GPT-4o-mini) and receives verifiable reward signals based on the patient's performance on the comprehension test. This two-stage alignment enables emergent instructional behaviors in SLMs.  
    \textbf{(Right: Evaluation)} We evaluate NoteAid-Chatbot with the Gold comprehension dataset and conduct general evaluation and a turing test.
    Above illustrates the generation evaluation based on the simulation with a virtual patient simulated from gold and silver dataset. We evaluated medical content generation and medical conversational strategies of our model. Below illustrates the  NoteAid-Chatbot in turing test. The NoteAid-Chatbot poses questions derived from a patient's discharge note to improve their understanding through interactive question answering. At the end of the session, the patient completes an exam assessing comprehension, which serves as the measurable learning outcome.
}
    \label{fig:overview}
\end{figure*}

% INTRODUCTION FLOW
% Talk about patient education and discharge scenario issues 
% 
% These issues can be sovled through chatbots
% The challenges : 
% i) data scarcity
% ii) domain-specific (reliability)
% iii) usability (lightweight, privacy preserving)
% OUR SOLUTION => CHATBOT NOTEAID

% patient and discharge 

Patients’ access to their electronic health record (EHR) notes, called OpenNotes~\cite{delbanco_inviting_2012}, represents a personalized communication channel. OpenNotes has been shown to enhance disease understanding~\cite{bronson_impact_1986}, patient–provider communication~\cite{homer_introduction_1999, elbourne_newbury_1987}, medication safety~\cite{assiri_impact_2022}, self-managed care~\cite{homer_introduction_1999}, and health outcomes~\cite{bronson_impact_1986, apter_home_2019}. A patient’s ability to comprehend EHRs is related to his/her level of health literacy, which is defined by the Institute of Medicine as “the degree to which individuals have the capacity to obtain, process, and understand basic information and services needed to make appropriate decisions regarding their health.”~\cite{institute_of_medicine_us_committee_on_health_literacy_health_2004} Thirty-six percent of American adults have limited health literacy~\cite{kutner2006health} and have shown difficulty in comprehending EHRs~\cite{pyper2004patients, keselman2007towards, chapman2003lay, lerner2000medical, jones1992patient, baldry1986giving}. Limited health literacy has been identified as one of the major barriers to patient portal use~\cite{sarkar2010literacy, zarcadoolas2013consumers, tieu2015barriers}.

\blfootnote{$\ddagger$ To appear in Findings of the Association for Computational Linguistics: EMNLP 2025}

Therefore, we developed NoteAid-Chatbot, a novel multi-component artificial intelligence (AI) system that helps patients comprehend their EHR notes. Communication is the central process of education~\cite{dewey_democracy_2004} In learning as conversation~\cite{sharples_learning_2005}, a patient does not read an EHR note, but gains information and knowledge through conversation with a Chatbot that reads the note.

One challenge for developing NoteAid-Chatbot is the lack of training data, making the development cost prohibitive. 
Therefore, inspired by \cite{deepseek-ai_deepseek-r1_2025}, we developed a novel training paradigm that is based on a multi-agent framework where we leverage state-of-the-art large language models (LLMs) and reinforcement learning (RL).
Our training framework is fully automated, eliminating the most of the costly human annotations for training. 
To help deploy our system to mobile devices~\cite{hutson2024forget,griewing2024proof,wang2024efficient,lu2024small}, we built upon LLaMA 3.2-3B-Instruct~\cite{dubey2024llama}, a lightweight open-source LLM. 
%Inspired by the promising results of initial RL-based alignment studies, 
We propose a two-stage training approach—initially applying supervised fine-tuning on a synthetic conversational dataset followed by simulated interactions between NoteAid-Chatbot and patient agent. 
We developed NoteAid-Chatbot using reinforcement learning, and our evaluation results by domain experts demonstrate that the basic Proximal policy optimization (PPO)~\cite{schulman_proximal_2017}, where the rewards are directly measured by patients' comprehension scores, achieved an excellent performance. 

\noindent\textbf{Our contributions are as follows:}
\begin{itemize}
    \item We propose an automated two stage multi-agent framework that produces a lightweight chatbot. ~\footnote{Our code and data is released at \url{https://github.com/memy85/2024_chatbot_noteaid} and \url{https://huggingface.co/datasets/bio-nlp-umass/NoteAid\_Chatbot} with CC-BY-NC 4.0 license}
    \item We successfully show that utilizing synthetic datasets can help in further steps of supervised fine-tuning and RL-based alignments resulting in a robust chatbot.
    \item We conduct a Turing test and showed the model is capable of educating humans better than non-experts. We also conduct an in-depth case studies and human-aligned evaluations to assess the NoteAid-Chatbot’s behavior in realistic, goal-oriented conversations.
\end{itemize}

\section{Development of NoteAid-Chatbot}

\subsection{Dataset Preparation and Configurations}
\begin{table}[]
     \caption{Demographic Category of synthetic dataset ${Comp}_S$}
     \label{tab:silver-note-demographic}
    \centering
    \resizebox{\linewidth}{!}{
    \begin{tabular}{lll}
    \toprule
         Category & Contents & Ratio\\
         \midrule
         Age & Young Adult (19--35 years) & 0.250 \\
          & Middle-aged Adult (36--55 years) & 0.350 \\
          & Older Adult (56--75 years) & 0.250\\
          & Elderly (76+ years) & 0.150 \\
         Gender & Male & 0.471 \\
                & Female &  0.529 \\
         Ethnicity & White & 0.672 \\
         & Black or African American & 0.100 \\
         & Hispanic or Latino & 0.100 \\
         & Asian & 0.080 \\
         & Native American or Alaska Native & 0.020 \\
         & Native Hawaiian or Pacific Islander & 0.015 \\
         & Mixed or Multicultural & 0.013 \\
    \bottomrule
    \end{tabular}
    }
\end{table}

We constructed two distinct comprehension datasets to assess patients' understanding of clinical notes: (i) a set of 100 real discharge notes, and (ii) a set of 10,000 synthetic discharge notes. We refer to these as the Gold and Silver datasets, denoted by $Comp_{G}$ and $Comp_{S}$, respectively.

The Gold dataset ($Comp_G$) comprises 50 discharge notes sampled from the MIMIC-IV database~\cite{johnson_mimic-iv_2023} and 50 notes obtained from the University of Pittsburgh Medical Center (UPMC) dataset which is a private dataset that cannot be disclosed. For each note, domain experts manually created between 5 and 10 multiple-choice questions and answers, denoted as $Q_{Gold}$. The instructions and the detailed procedure for QA generation are included in the appendix~\ref{appendix:dataset_evaluation}.
%to assess patient understanding or comprehension of the note.

The Silver dataset (${Comp}_S$) consists of synthetically generated discharge notes (Table~\ref{tab:silver-note-demographic}). For each note, we used GPT-4o-mini to create the comprehension QA $Q_S$. The prompts used for note and QA generation are also included in the appendix~\ref{appendix}.
%The prompt we used for data generation is included in the supplementary section.
In addition, for each note, we also generated a simulated conversation history between an educator agent and a patient agent. We define this conversation dataset as ${Conv}_{S}$, We used ${Conv}_{S}$ to supervise fine-tune NoteAid-Chatbot and then we deployed reinforcement learning based on how well the model achieved the comprehension scores on $Q_S$. We used the conversation data generated from 8000 notes for supervised finetuning. The reinforcement learning was trained on the comprehension QA dataset created from the remaining 2000 notes.   
%denoted as $\mathcal{C}_{S}$, also a discharge questionnaire $Q_S$ using the GPT-4o-mini model. 
Detailed procedures for data generation and evaluation are provided in the appendix~\ref{appendix:dataset_evaluation}.

For Gold and Silver dataset of notes, we formally note as : 

\begin{equation}
{Comp}_{G} = \{(\mathcal{N}^i_{G}, Q_G^i) | i \in \text{[1,100]} \}
\end{equation}
\begin{equation}
{Comp}_{S} = \{(\mathcal{N}^{i}_{S}, {Conv}_{S}^{i}, Q_S^i) | i \in \text{[1,10000]} \}
\end{equation}

\subsection{Supervised Fine-Tuning stage}
%As proposed in DeepSeek-R1, 

% Note that we use synthetic dataset $\mathcal{N}_{S}$ and the generated conversation history $\mathcal{C}_{S}$ to train our model. 
We first trained the open-source LLaMA 3.2-3B-Instruct on 80\% of portion of Silver dataset ${Comp}_{S}$. We employed Low Rank Adapation (LoRA)~\cite{hu_lora_2021} to fine-tune the model, and report the result. 
We insert the $\mathcal{N}_S$ to the system prompt and instruction fine-tuned on ${Conv}_S$. With the synthetic dataset's quality well controlled, we can enable the model to be trained on domain-specific tasks.

\subsection{Reinforcement Learning stage}

\begin{figure}[h]
    \centering
    \includegraphics[width=\linewidth, trim=2cm 0cm 1.9cm 0cm,clip]{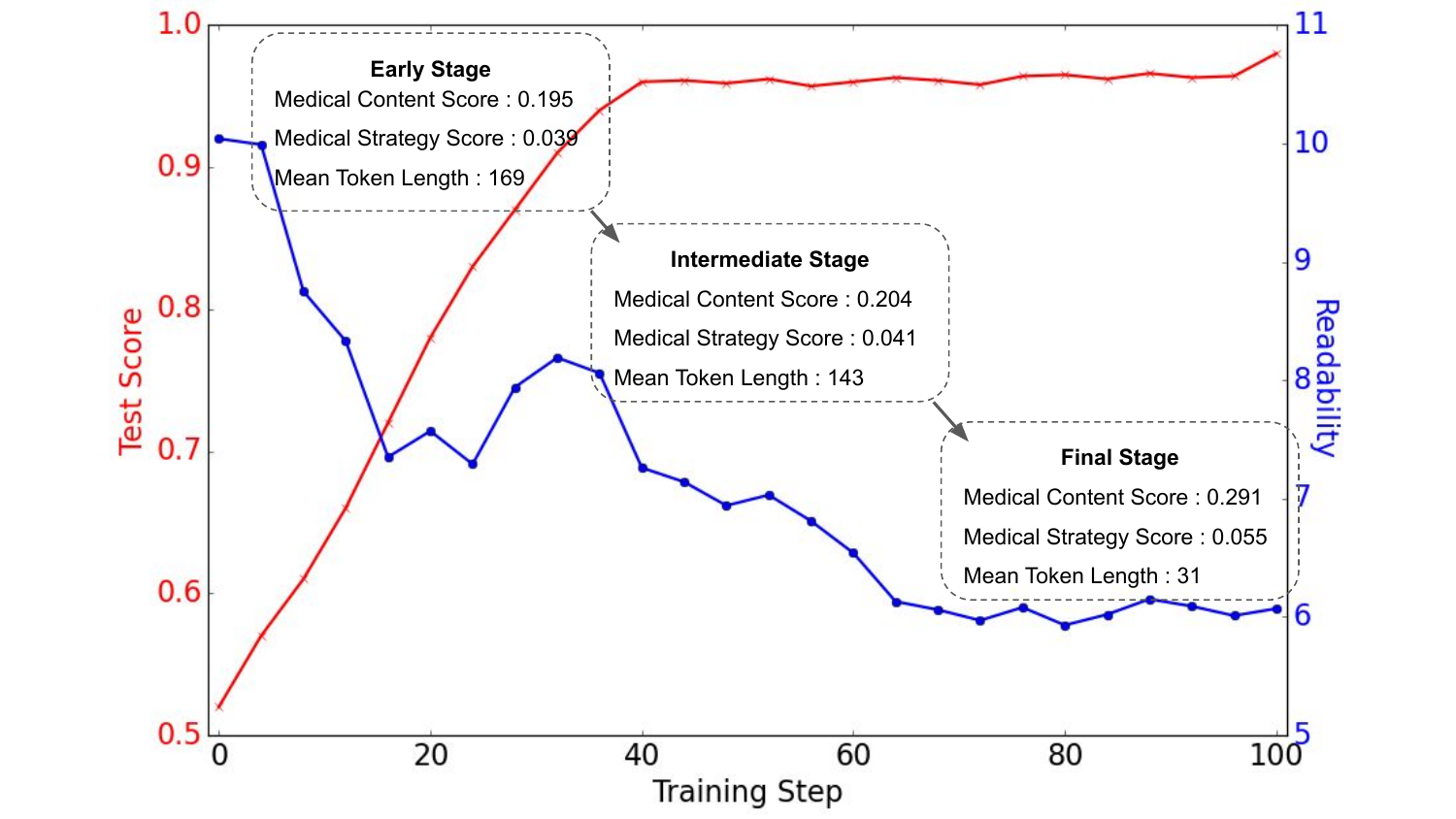}
    \caption{As the RL-based alignment training progress, the comprehension score increases while the FKGL score of the text decreases. We also see an increase in  Medical Content score, Medical conversation strategy score. While the mean token length generated decreases in each training steps during reinforcement learning stage.}
    \label{fig:alignment}
\end{figure}
Following the supervised fine-tuning stage, we further optimize NoteAid-Chatbot within a Reinforcement Learning (RL) framework. In this stage, we simulate a dialogue scenario in which the fine-tuned model assumes the role of an educator interacting with a patient in a discharge context. To simulate patient behavior, we utilize both $\mathcal{N}_{S}$ scenarios and the GPT-4o-mini model as the patient agent.

For each simulation, the dialogue is initiated using a specific discharge note $\mathcal{N}_{S}^i$. The educator agent (NoteAid-Chatbot) is tasked with conveying key information from the discharge note to the patient agent. Upon completion of the interaction, the patient agent is assessed using the corresponding set of questions $Q_{S}^i$ we created for the note. The patient’s performance on this assessment is then used to compute the reward signal for the reinforcement learning update. We reward the model using reward function in ~\ref{appendix:reward-function}

% Note that our framework does use $\mathcal{D}_{Gold}$ in its alignment. However, we show that even for $\mathcal{D}_{Silver}$ and in generation metrics along with human evaluation and thorough case studies, that the pipeline is reliable.  

\section{Chatbot Evaluation}

\subsection{Chatbot evaluation}
We evaluated NoteAid-Chatbot with four distinct measures. (i) Generation metrics, (ii) Medical contents, (iii) Medical conversation strategy and (iv) Human evaluation.

\subsubsection{Generation Metric}
We evaluated generation quality using a set of established metrics, including BLEU~\cite{papineni_bleu_2002}, ROUGE-L~\cite{lin_rouge_2004}, BERTScore~\cite{zhang_bertscore_2020}, and the Flesch-Kincaid Grade Level (FKGL)~\cite{flesch_flesch-kincaid_2007}. For this evaluation, we employed a held-out subset of $\mathcal{N}_S$ and ${Conv}_{S}$ that were not seen during the chatbot’s training phase. We simulate a conversation between the educator and the patient, and evaluate the educator's utterance based on ${Conv}_S$. BLEU, ROUGE-L, and BERTScore assess the semantic alignment between the model-generated utterances and the reference texts in the test set. The FKGL or the readability score, quantifies the ease with which the generated text can be understood by human readers. Lower FKGL indicates it is easier for the reader to understand the text, which means higher readability, and vice versa.

\subsubsection{Medical contents and Conversation strategies}

The model’s conversational ability was evaluated using the criteria shown in Table~\ref{tab1-medcontent-strategy}. Dialogues were simulated between the chatbot and an AI patient implemented with GPT-4o-mini. To ensure consistency and prevent excessively long interactions, a maximum of 20 dialogue turns was imposed. Evaluation followed the LLM-as-a-judge framework, in which GPT-4o-mini assessed the quality of the conversation history based on predefined prompts.

To measure the chatbot’s capacity for domain-specific dialogue, two evaluation criteria were established (Table~\ref{tab1-medcontent-strategy}). The first criterion examined whether the model could effectively communicate medically relevant content within the context of discharge scenarios. Following the framework proposed by \citet{desai_empowering_2021}, the chatbot was expected to address six key informational categories. For each utterance ($h_i$) within the chatbot’s conversation history ($H$), we identified the presence of each content category ($C_k$) and computed a normalized score by dividing the count by the total token length of the utterance, as defined in Equation~\ref{eq:contentscore}

\begin{equation}
\label{eq:contentscore}
% ContentScore(C_k, H) = 
\text{Content Score}=\frac{1}{m} \sum^m_{i=1}  \frac{Count(C_k, h_i)}{\log NumToken(h_i)}
\end{equation}

% Since the chatbot has to interact with the patient, we evaluate the model's conversational strategies, using the criteria suggested from \citet{king_best_2013}. This criteria measures whether the educator communicated in the ideal manner that clinicians communicate with patients. We asked GPT-4o-mini to evaluate the conversation history to score ($S_k$) each category in a 1 to 5 likert scale and also normalized them with the total token length of the utterance shown in formula~\ref{eq:strategyscore}. 

To evaluate the chatbot's conversational strategies, we adopt the criteria proposed by \citet{king_best_2013}, which assess whether the communication aligns with ideal clinician-patient interactions. Using GPT-4o-mini as an evaluator, we scored each category ($S_k$) on a 1–5 Likert scale and normalized the scores by the total token length of the corresponding utterances, as shown in formula~\ref{eq:strategyscore}.

\begin{equation}
\label{eq:strategyscore}
% StrategyScore(S_k, H) =  
\text{Strategy Score}=\frac{\text{LLM-as-a-Judge}(S_k, H)}{\log NumToken(H)}
\end{equation}

\begin{table}[h]
  \caption{Medical content and conversation strategy criterias}
  \begin{adjustbox}{width=0.9\columnwidth, center}
  \label{tab1-medcontent-strategy}
  \centering
  \begin{tabular}{llll}
    \toprule
    \textbf{Medical Contents}    \\ 
    \cite{desai_empowering_2021} \\
    \midrule
    Return to the Hospital/Emergency Department \\
    Medication  \\
    Discharge Diagnosis       \\
    Post-discharge treatment  \\
    Test and treatments during stay \\
    Follow-up \\
    \midrule
    \textbf{Medical Conversation Strategies}  \\
    \cite{king_best_2013} \\
    \midrule
    Fostering relationship     \\
    Gathering information      \\
    Providing information  \\
    Decision making \\
    Responding to emotions \\
    Enabling disease and treatment-related behavior \\
    \bottomrule
  \end{tabular}
\end{adjustbox}
\end{table}

 To ensure the quality of the evaluation, we conduct a case study for the conversation histories and the output of the evaluation to show that the LLM-as-a-judge evaluation results are reliable and acceptable. 

\subsubsection{Turing test}

% \begin{table}
%   \caption{Usability survey contents}
  
%  \begin{adjustbox}{width=\linewidth, center}
%   \label{tab-human-eval-survey-contents}
%   \centering
%   \begin{tabular}{llll}
  
%     \toprule
%     \textbf{Index} & \textbf{Questions}  & \textbf{Category} & \textbf{Responses} \\
%     \midrule
%     1 & Interacting with the educator felt smooth. &  Usability &  1-5\\
%     2 & The educator's conversation was consistent throughout the interaction &  Consistency.  &  1-5 \\
%     3 & The educator responded in a way that felt natural and human-like. &  Realness  & 1-5 \\
%     4 & I trusted the information provided by the educator. &  Trustworthiness & 1-5 \\
%     5 & The educator was able to understand and address my concerns. & Perceived understanding & 1-5 \\
%     6 & The educator helped me better understand my discharge instructions. &  Usefulness & 1-5 \\
%     7 & I felt satisfied with the overall experience with the educator. &  Satisfaction & 1-5 \\
%     \midrule
%     8 & Overall, did the educator feel like a human? &  Realness & Yes/No/Not Sure\\
%     \bottomrule
%   \end{tabular}
%   \end{adjustbox}
% \end{table}

To further evaluate the usability and effectiveness of the chatbot, we conducted a Turing test involving three experimental groups: Group A (non-expert–patient interaction), Group B (expert–patient interaction), and Group C (chatbot–patient interaction). Participants assigned to the non-expert and patient roles were recruited from the university. For each patient, we assigned a note from $\mathcal{N}_G$, where the educating side can only see the discharge note and the patients cannot. Additional details regarding the experimental setup are provided in the appendix ~\ref{appendix:human-evaluation}.

Each participant assigned to the patient role first completed a health literacy assessment (S-TOFHLA test; Short version of Test Of Functional Health Literacy in Adults;TOFHLA) to establish a baseline~\cite{parker_test_1995}. Following this, participants engaged in a 15-minute educational session conducted via a chat-based interface. During the session, only the educator (i.e., the human or chatbot in Groups A–C) had access to the corresponding discharge note, while the patient engaged in dialogue to learn about the content. The patients were not disclosed about the identity of their educator until the experiment was finished.  

Upon completion of the session, patients were administered a comprehension test from $Q_G$ based on the discharge note, assessing their interaction experience. The outcomes of the comprehension and usability measures were then analyzed across the three groups to assess the relative effectiveness of each educator type. This human subjects research was approved by the university IRB. All participant consents were obtained before they began the study, they were given 10 dollar worth of gift card for their participation as a compensation.  

\subsection{Baseline models}

We compared our chatbot with closed- and open-source LLMs. For the closed-source models we used GPT-4o-mini and GPT-4.1\footnote{https://openai.com/index/gpt-4-1/} from~\citet{openai_website}. For the open-source model we used LLaMA 3.2-3B-Instruct that were supervised fine-tuned on ${Comp}_{S}$ train set with LoRA technique. We also added experiments using BioMistral 7B~\cite{labrak_biomistral_2024} which is a LLM pretrained on medical texts and Qwen3-8B~\cite{yang_qwen3_2025} model which is known for its strong reasoning capabilites.

\section{Experimental results}

\subsection{Basic evaluation}

% \begin{table}
%   \caption{Generation metric evaluation on ${Conv}_S \in {Comp}_{S}$ test set. NoteAid-Chatbot (LLaMA3.2-3B+ LoRA + PPO) showed the higher scores in every metric compared to baseline models.}
% \begin{adjustbox}{width=0.99\linewidth, center}
%   \label{tab:generation-metric-results}
%   \centering
%   \begin{tabular}{lcccc}
%     % \cmidrule(r){1-2}
%     \toprule
%     \textbf{Model} & \textbf{BLEU} & \textbf{ROUGE-L} & \textbf{BERTscore} & \textbf{Readability} \\
%     \midrule
%     GPT-4o-mini &  0.020  &   0.119    & 0.853   & 10.672  \\
%     GPT-4.1 &  0.014  &   0.102 & 0.853   & 10.947  \\
%     \midrule
%     LLaMA3.2-3B &  0.023  &   0.112 & 0.851  & 10.777  \\
%      % + SFT &  0.025  &   0.104 & 0.830 $\pm$ \text{\small 0.001} & 7.905 $\pm$ \text{\small 0.207} \\
%      + LoRA &  0.031 &   0.125 & 0.851 & 7.636  \\
%      + LoRA + PPO &  \textbf{0.157} &   \textbf{0.322} & \textbf{0.893} & \textbf{7.237}  \\
%      % + LoRA + GRPO &  0.140 &   0.292 & 0.889 & 7.202 \\
%     \bottomrule
%   \end{tabular}
% \end{adjustbox}
% \end{table}
\begin{table}
  \caption{Generation metric evaluation on ${Conv}_S \in {Comp}_{S}$ test set. NoteAid-Chatbot (LLaMA3.2-3B+ LoRA + PPO) showed the higher scores in every metric compared to baseline models.}
\begin{adjustbox}{width=0.99\linewidth, center}
  \label{tab-generation-metric-results}
  \centering
  \begin{tabular}{lcccc}
    % \cmidrule(r){1-2}
    \toprule
    \textbf{Model} & \textbf{BLEU} & \textbf{ROUGE-L} & \textbf{BERTscore} & \textbf{Readability} \\
    \midrule
    % GPT-4o-mini &  0.023  &   0.119    & 0.853 $\pm$ \text{\small 0.001}  & 10.672 $\pm$ \text{\small 0.100} \\
    % GPT-4.1 &  0.014  &   0.102 & 0.853 $\pm$ \text{\small 0.001}  & 10.947 $\pm$ \text{\small 0.140} \\
    % \midrule
    GPT-4o-mini & 0.023 $\pm$ \text{\small 0.002} & 0.128 $\pm$ \text{\small 0.007} & 0.853 $\pm$ \text{\small 0.001} & 10.672 $\pm$ \text{\small 0.100} \\
    GPT-4.1     & 0.014 $\pm$ \text{\small 0.001} & 0.108 $\pm$ \text{\small 0.005} & 0.853 $\pm$ \text{\small 0.001} & 10.947 $\pm$ \text{\small 0.140} \\
    BioMistral 7B & 0.030 $\pm$ \text{\small 0.007} & 0.137 $\pm$ \text{\small 0.020} & 0.770 $\pm$ \text{\small 0.010} & 26.077 $\pm$ \text{\small 2.680} \\
    Qwen3-8B & 0.014 $\pm$ \text{\small 0.001} & 0.115 $\pm$ \text{\small 0.006} & 0.835 $\pm$ \text{\small 0.001} & 11.371 $\pm$ \text{\small 0.163} \\
    \midrule
    LLaMA3.2-3B        & 0.024 $\pm$ \text{\small 0.002} & 0.118 $\pm$ \text{\small 0.006} & 0.851 $\pm$ \text{\small 0.001} & 10.777 $\pm$ \text{\small 0.230} \\
    + LoRA             & 0.033 $\pm$ \text{\small 0.003} & 0.127 $\pm$ \text{\small 0.008} & 0.851 $\pm$ \text{\small 0.001} & 7.636 $\pm$ \text{\small 0.086} \\
    + LoRA + PPO       & \textbf{0.153} $\pm$ \text{\small 0.017} & \textbf{0.325} $\pm$ \text{\small 0.021} & \textbf{0.893} $\pm$ \text{\small 0.002} & \textbf{7.237} $\pm$ \text{\small 0.134} \\
    % LLaMA3.2-3B &  0.023  &   0.112 & 0.851 $\pm$ \text{\small 0.001} & 10.777 $\pm$ \text{\small 0.230} \\
    %  % + SFT &  0.025  &   0.104 & 0.830 $\pm$ \text{\small 0.001} & 7.905 $\pm$ \text{\small 0.207} \\
    %  + LoRA &  0.031 &   0.125 & 0.851 $\pm$ \text{\small 0.001}& 7.636 $\pm$ \text{\small 0.086} \\
    %  + LoRA + PPO &  \textbf{0.157} &   \textbf{0.322} & \textbf{0.893} $\pm$ \text{\small 0.002}& \textbf{7.237} $\pm$ \text{\small 0.134} \\
    %  % + LoRA + GRPO &  0.140 &   0.292 & 0.889 & 7.202 \\
    \bottomrule
  \end{tabular}
\end{adjustbox}
\end{table}

Our NoteAid-Chatbot (LLaMA3.2-3B with LoRA and PPO) showed the highest performance in every metrics that were measured in Table~\ref{tab-generation-metric-results}. This shows that NoteAid-Chatbot's ability to align with ${Comp}_{S}$ was even more enhanced than fine-tuned model (LLaMA 3.2-3B + LoRA) achieving 0.157, 0.322, 0.893 and 7.237 for BLEU, ROUGE-L, BERTscore and Readability respectively. Remember that in the supervised fine-tuning stage, the model was trained on ${Comp}_{S}$ train set. Fine-tuning can enhance the BLEU, ROUGE-L or BERTscore, but the enhancements were trivial compared to the enhancements of using reinforcement learning, while NoteAid-Chatbot achieves superiority compared to other baselines. Our chatbots were able to capture the semantics using reinforcement learning. Also, note that the FKGL scores are lower in our chatbot, implying that the texts are much easier to read. Trained models shows lower score of grade level which means the texts are easier to read.

\subsection{Medical contents and Conversation strategy evaluation}

\begin{table*}[h!]
\caption{Medical content evaluation on $\mathcal{N}_{G}$ (top) and $\mathcal{N}_{S}$ (bottom). We find that NoteAid-Chatbot (LoRA + PPO) successfully covers the core medical topics in the discharge scenario with more efficiency. The scores are calculated using Eq.~\ref{eq:contentscore}}
\begin{adjustbox}{width=\linewidth, center}
  \label{tab:content-gold-silver}
  \centering
  \begin{tabular}{lcccccc}
    % \cmidrule(r){1-2}
    \toprule
    \textbf{Model} & \textbf{Diagnosis} & \textbf{Follow-up} & \textbf{Medication} & \textbf{Post-discharge treatment}  & \textbf{Return to Hospital/ED} & \textbf{Tests/Treatments}\\ 
    \midrule
    GPT-4o-mini & $0.241 \pm \text{\small 0.003}$ & $0.232 \pm \text{\small 0.002}$ & $0.231 \pm \text{\small 0.002}$ & $0.232 \pm \text{\small 0.002}$ & $0.234 \pm \text{\small 0.002}$ & $0.232 \pm \text{\small 0.004}$ \\
GPT-4.1 & $0.193 \pm \text{\small 0.001}$ & $0.219 \pm \text{\small 0.002}$ & $0.214 \pm \text{\small 0.002}$ & $0.212 \pm \text{\small 0.002}$ & $0.218 \pm \text{\small 0.002}$ & $0.197 \pm \text{\small 0.003}$ \\
BioMistral 7B & $0.239 \pm \text{\small 0.012}$ & $0.240 \pm \text{\small 0.006}$ & $0.233 \pm \text{\small 0.006}$ & $0.233 \pm \text{\small 0.005}$ & $0.235 \pm \text{\small 0.006}$ & $0.227 \pm \text{\small 0.008}$ \\ 
Qwen3-8B & $0.233 \pm \text{\small 0.004}$ & $0.236 \pm \text{\small 0.003}$ & $0.232 \pm \text{\small 0.004}$ & $0.233 \pm \text{\small 0.003}$ & $0.236 \pm \text{\small 0.003}$ & $0.233 \pm \text{\small 0.005}$ \\
LLaMA3.2-3B & $0.196 \pm \text{\small 0.003}$ & $0.204 \pm \text{\small 0.003}$ & $0.199 \pm \text{\small 0.002}$ & $0.197 \pm \text{\small 0.002}$ & $0.201 \pm \text{\small 0.002}$ & $0.197 \pm \text{\small 0.004}$ \\
% + SFT & $0.193 \pm \text{\small 0.002}$ & $0.201 \pm \text{\small 0.003}$ & $0.199 \pm \text{\small 0.002}$ & $0.197 \pm \text{\small 0.002}$ & $0.200 \pm \text{\small 0.002}$ & $0.197 \pm \text{\small 0.003}$ \\
+ LoRA & $0.209 \pm \text{\small 0.004}$ & $0.222 \pm \text{\small 0.003}$ & $0.219 \pm \text{\small 0.003}$ & $0.220 \pm \text{\small 0.003}$ & $0.221 \pm \text{\small 0.003}$ & $0.211 \pm \text{\small 0.006}$ \\
+ LoRA + PPO & $\mathbf{0.287} \pm \text{\small 0.004}$ & $\mathbf{0.286} \pm \text{\small 0.003}$ & $\mathbf{0.292} \pm \text{\small 0.004}$ & $\mathbf{0.294} \pm \text{\small 0.005}$ & $\mathbf{0.301} \pm \text{\small 0.004}$ & $\mathbf{0.286} \pm \text{\small 0.005}$ \\
% + LoRA + GRPO & \mathbf{0.297} \\
    
    \midrule
    GPT-4o-mini & $0.247 \pm \text{\small 0.002}$ & $0.235 \pm \text{\small 0.002}$ & $0.233 \pm \text{\small 0.002}$ & $0.234 \pm \text{\small 0.002}$ & $0.236 \pm \text{\small 0.002}$ & $0.236 \pm \text{\small 0.003}$ \\
GPT-4.1 & $0.197 \pm \text{\small 0.002}$ & $0.217 \pm \text{\small 0.002}$ & $0.211 \pm \text{\small 0.002}$ & $0.212 \pm \text{\small 0.002}$ & $0.216 \pm \text{\small 0.002}$ & $0.200 \pm \text{\small 0.003}$ \\
LLaMA3.2-3B & $0.199 \pm \text{\small 0.004}$ & $0.204 \pm \text{\small 0.003}$ & $0.200 \pm \text{\small 0.002}$ & $0.199 \pm \text{\small 0.002}$ & $0.201 \pm \text{\small 0.002}$ & $0.198 \pm \text{\small 0.004}$ \\
% + SFT & $0.194 \pm \text{\small 0.002}$ & $0.201 \pm \text{\small 0.002}$ & $0.199 \pm \text{\small 0.002}$ & $0.197 \pm \text{\small 0.001}$ & $0.200 \pm \text{\small 0.002}$ & $0.198 \pm \text{\small 0.003}$ \\
BioMistral 7B & $0.221 \pm \text{\small 0.006}$ & $0.230 \pm \text{\small 0.006}$ & $0.227 \pm \text{\small 0.005}$ & $0.227 \pm \text{\small 0.005}$ & $0.229 \pm \text{\small 0.005}$ & $0.225 \pm \text{\small 0.007}$ \\ 
Qwen3-8B & $0.234 \pm \text{\small 0.003}$ & $0.241 \pm \text{\small 0.002}$ & $0.234 \pm \text{\small 0.002}$ & $0.236 \pm \text{\small 0.002}$ & $0.239 \pm \text{\small 0.002}$ & $0.234 \pm \text{\small 0.005}$ \\
+ LoRA & $0.206 \pm \text{\small 0.004}$ & $0.223 \pm \text{\small 0.003}$ & $0.219 \pm \text{\small 0.003}$ & $0.222 \pm \text{\small 0.003}$ & $0.221 \pm \text{\small 0.003}$ & $0.214 \pm \text{\small 0.008}$ \\
+ LoRA + PPO & $\mathbf{0.285} \pm \text{\small 0.005}$ & $\mathbf{0.280} \pm \text{\small 0.002}$ & $\mathbf{0.287} \pm \text{\small 0.003}$ & $\mathbf{0.293} \pm \text{\small 0.005}$ & $\mathbf{0.301} \pm \text{\small 0.005}$ & $\mathbf{0.286} \pm \text{\small 0.004}$ \\
    \bottomrule
  \end{tabular}
\end{adjustbox}
\end{table*}

In the evaluation of medical content, NoteAid-Chatbot demonstrated the ability to effectively cover the essential topics typically addressed in conversations between educators and patients (Table~\ref{tab:content-gold-silver}). In every aspect, our RL-based alignment showed superior performance compared to the baseline models. Reinforcement learning contributed to more concise utterances by reducing the number of generated tokens while preserving the relevance and completeness of the conveyed information. As the patient-side questionnaires were designed based on the content framework proposed by \cite{desai_empowering_2021}, the alignment between the model's outputs and the expected content was further reinforced. This alignment allows the model to deliver critical information more efficiently, outperforming baseline models in both content coverage and token economy.

\begin{table*}[]
\caption{Medical conversation strategy evaluation on $\mathcal{N}_{G}$ (top) and $\mathcal{N}_{S}$ (bottom). As illustrated, NoteAid-Chatbot (+ LoRA + PPO) successfully uses the core strategies that are recommended for medical conversations in the discharge scenario. The scores are calculated using Eq.~\ref{eq:strategyscore}}
\begin{adjustbox}{width=0.95\linewidth, center}
  \label{tab:strategy-gold-silver}
  \centering
  \begin{tabular}{lcccccc}
    % \cmidrule(r){1-2}
    \toprule
    \textbf{Model} & \textbf{Fostering } & \textbf{Gathering} & \textbf{Providing} & \textbf{Decision}  & \textbf{Enabling disease} & \textbf{Responding to}\\ 
    & \textbf{relationship} & \textbf{information} & \textbf{information} & \textbf{making} & \textbf{and treatment-related} & \textbf{emotions} \\
    & & & & & \textbf{behavior} & \\
    \midrule
     GPT-4o-mini & $0.046 \pm \text{\small 0.000}$ & $0.046 \pm \text{\small 0.000}$ & $0.056 \pm \text{\small 0.001}$ & $0.044 \pm \text{\small 0.001}$ & $0.047 \pm \text{\small 0.001}$ & $0.038 \pm \text{\small 0.001}$ \\
GPT-4.1 & $0.043 \pm \text{\small 0.000}$ & $0.043 \pm \text{\small 0.000}$ & $0.053 \pm \text{\small 0.001}$ & $0.041 \pm \text{\small 0.001}$ & $0.043 \pm \text{\small 0.001}$ & $0.039 \pm \text{\small 0.001}$ \\
BioMistral 7B & $0.040 \pm \text{\small 0.002}$ & $0.032 \pm \text{\small 0.002}$ & $0.041 \pm \text{\small 0.003}$ & $0.031 \pm \text{\small 0.002}$ & $0.035 \pm \text{\small 0.002}$ & $0.030 \pm \text{\small 0.002}$ \\
Qwen3-8B & $0.049 \pm \text{\small 0.001}$ & $0.038 \pm \text{\small 0.001}$ & $0.058 \pm \text{\small 0.001}$ & $0.041 \pm \text{\small 0.001}$ & $0.051 \pm \text{\small 0.001}$ & $0.039 \pm \text{\small 0.002}$ \\
LLaMA3.2-3B & $0.040 \pm \text{\small 0.000}$ & $0.040 \pm \text{\small 0.001}$ & $0.048 \pm \text{\small 0.001}$ & $0.036 \pm \text{\small 0.001}$ & $0.039 \pm \text{\small 0.001}$ & $0.034 \pm \text{\small 0.001}$ \\
% + SFT & $0.040 \pm \text{\small 0.001}$ & $0.040 \pm \text{\small 0.001}$ & $0.048 \pm \text{\small 0.001}$ & $0.035 \pm \text{\small 0.001}$ & $0.039 \pm \text{\small 0.001}$ & $0.033 \pm \text{\small 0.001}$ \\
+ LoRA & $0.044 \pm \text{\small 0.001}$ & $0.044 \pm \text{\small 0.001}$ & $0.053 \pm \text{\small 0.001}$ & $0.041 \pm \text{\small 0.001}$ & $0.045 \pm \text{\small 0.001}$ & $0.038 \pm \text{\small 0.001}$ \\
+ LoRA + PPO & $\mathbf{0.059} \pm \text{\small 0.001}$ & $\mathbf{0.056} \pm \text{\small 0.002}$ & $\mathbf{0.061} \pm \text{\small 0.001}$ & $\mathbf{0.047} \pm \text{\small 0.001}$ & $\mathbf{0.058} \pm \text{\small 0.001}$ & $\mathbf{0.046} \pm \text{\small 0.001}$ \\
    \midrule
     GPT-4o-mini & $0.046 \pm \text{\small 0.000}$ & $0.047 \pm \text{\small 0.001}$ & $0.057 \pm \text{\small 0.001}$ & $0.045 \pm \text{\small 0.001}$ & $0.047 \pm \text{\small 0.001}$ & $0.040 \pm \text{\small 0.001}$ \\
GPT-4.1 & $0.043 \pm \text{\small 0.000}$ & $0.043 \pm \text{\small 0.000}$ & $0.053 \pm \text{\small 0.001}$ & $0.041 \pm \text{\small 0.001}$ & $0.044 \pm \text{\small 0.001}$ & $0.039 \pm \text{\small 0.001}$ \\
BioMistral 7B & $0.045 \pm \text{\small 0.002}$ & $0.043 \pm \text{\small 0.002}$ & $0.049 \pm \text{\small 0.002}$ & $0.037 \pm \text{\small 0.002}$ & $0.043 \pm \text{\small 0.002}$ & $0.036 \pm \text{\small 0.002}$ \\ 
Qwen3-8B & $0.047 \pm \text{\small 0.000}$ & $0.047 \pm \text{\small 0.001}$ & $0.057 \pm \text{\small 0.001}$ & $0.041 \pm \text{\small 0.001}$ & $0.049 \pm \text{\small 0.001}$ & $0.038 \pm \text{\small 0.001}$ \\
LLaMA3.2-3B & $0.040 \pm \text{\small 0.000}$ & $0.040 \pm \text{\small 0.001}$ & $0.049 \pm \text{\small 0.001}$ & $0.037 \pm \text{\small 0.001}$ & $0.041 \pm \text{\small 0.001}$ & $0.034 \pm \text{\small 0.001}$ \\
% + SFT & $0.040 \pm \text{\small 0.000}$ & $0.040 \pm \text{\small 0.000}$ & $0.048 \pm \text{\small 0.001}$ & $0.037 \pm \text{\small 0.001}$ & $0.040 \pm \text{\small 0.001}$ & $0.034 \pm \text{\small 0.001}$ \\
+ LoRA & $0.045 \pm \text{\small 0.001}$ & $0.044 \pm \text{\small 0.001}$ & $0.054 \pm \text{\small 0.001}$ & $0.042 \pm \text{\small 0.001}$ & $0.045 \pm \text{\small 0.001}$ & $0.037 \pm \text{\small 0.001}$ \\
+ LoRA + PPO & $\mathbf{0.059} \pm \text{\small 0.001}$ & $\mathbf{0.056} \pm \text{\small 0.001}$ & $\mathbf{0.063} \pm \text{\small 0.001}$ & $\mathbf{0.048} \pm \text{\small 0.001}$ & $\mathbf{0.059} \pm \text{\small 0.001}$ & $\mathbf{0.046} \pm \text{\small 0.001}$ \\
    \bottomrule
  \end{tabular}
\end{adjustbox}
\end{table*}

In terms of medical strategy adherence, our Chatbot is capable of producing concise responses while still aligning with established medical communication guidelines (Table~\ref{tab:strategy-gold-silver}). Although explicit instructions or reward signals for conversational strategies were not incorporated during reinforcement learning, some degradation of these traits was observed over the course of training. Nevertheless, due to the initial supervised fine-tuning on datasets explicitly designed to model such strategies, the model retains several key characteristics of effective medical dialogue. These results suggest that, with a balanced training regimen, it is possible to preserve conversational quality that aligns with the criteria outlined in Table~\ref{tab1-medcontent-strategy}.

\subsection{LLM-as-a-Judge Evaluation Case studies}

We performed a specific case study to ensure the quality of the LLM-as-a-judge results and also validate the effectiveness of our NoteAid-Chatbot. As seen in Table~\ref{tab:content-case} in appendix~\ref{appendix:medical-content-evaluation}, GPT-4o-mini has classified the utterance of the NoteAid-Chatbot based on the criterias suggested in Table~\ref{tab1-medcontent-strategy}. We observed that the model successfully classifies the utterances of the NoteAid-Chatbot with a high precision. Since it is possible that the more the NoteAid-Chatbot generates tokens, it is likely to cover the medical contents that should be addressed, we normalize the counts of the categories with the length of the utterance as seen in formula~\ref{eq:contentscore}. Grounded on Table~\ref{tab:content-gold-silver} and the case studies from Table~\ref{tab:content-case}, we can see that NoteAid-Chatbot generates less tokens but successfully covers the details that needs to be covered in the conversation simulation.

% In appendix~\ref{appendix:medical-conversation-strategy}, we also show some results from the LLM-as-a-judge for medical conversation strategy results. 
Based on the conversation history between the patient agent and our chatbot, GPT-4o-mini will evaluate the conversational strategy scores. We found that the model's response has little discrepancy with expert annotators (Appendix~\ref{appendix:medical-conversation-strategy}). Grounded on the evidence of ~\cite{zheng_judging_2023, tu_towards_2025, cai2023paniniqa}, GPT-4's judgements are highly aligned with human level evaluation. As seen in appendix Table~\ref{tab:case-medical-strategy}, our Chatbot still achieves these categories above 3 over 5 in overall evaluation which shows that the model still maintains the ideal conversational strategies after reinforcement learning stage. Also comparing the scores with other models in Table~\ref{tab:strategy-gold-silver}, NoteAid-Chatbot
achieves the highest scores.

It is important to note that the reinforcement learning alignment phase did not incorporate any explicit mechanisms for training conversational strategies. Instead, such strategies were derived from the dataset and acquired during the supervised fine-tuning stage. This indicates that knowledge and communicative behaviors learned through fine-tuning can be preserved throughout subsequent reinforcement learning. Investigating optimal combinations and interactions between supervised fine-tuning and reinforcement learning represents a promising direction for future research.

\subsection{Alignment through Multi-Agent Framework}

As shown in Figure~\ref{fig:alignment} reinforcement learning for alignment substantially enhances the chatbot’s performance and response quality. As seen in Figure ~\ref{fig:response-comparison}, for the same question, the model generates a shorter version of the response. But note that there are minimal loss of information and the model successfully delivers the core contents that the patient should know in a polite and simplified manner. \cite{ouyang_training_2022, yang_aligning_2025, zhang_med-rlvr_2025} suggests that reinforcement learning can effectively address limitations of supervised fine-tuning—where performance may stagnate or even deteriorate due to misalignment with desired behaviors. 
As illustrated in Figure~\ref{fig:alignment}, the model keeps learning to talk briefly as it simulates a conversation with the patient agent. Even with a simple reward framework, such as PPO, the model can enhance its conversational skills. Over the course of training, the model learns to produce more concise utterances. This brevity is advantageous, as longer outputs have a higher risk of introducing confusion, thereby impairing the patient agent’s ability to respond accurately. The emergence of shorter, clearer utterances is particularly valuable in the context of patient education, where materials are recommended to be written at or below a sixth- to eighth-grade reading level~\cite{okuhara_readability_2025, stossel_readability_2012}. Failure to meet this standard can significantly hinder patient comprehension of discharge instructions. One explanation of this phenomenon is that over the course of RL-based alignment, the model learns to speak briefly because if it generates too much tokens, which has unnecessary information, it will decrease the score the patient agent gets correct. Our experimental results demonstrate that a carefully configured multi-agent reinforcement learning framework can yield a well-aligned, effective chatbot for healthcare communication.
Albeit our framework uses a simple, yet, indirect reward which relies on the performance of the patient agent, studies such as \cite{xu_is_2024}, demonstrates PPO-based methods shows robust effectiveness in larger scopes of tasks.

\subsection{Comparing Responses}

\begin{figure}[h]
    \centering
    \includegraphics[width=\columnwidth, trim=5cm 0cm 6cm 0cm, clip]{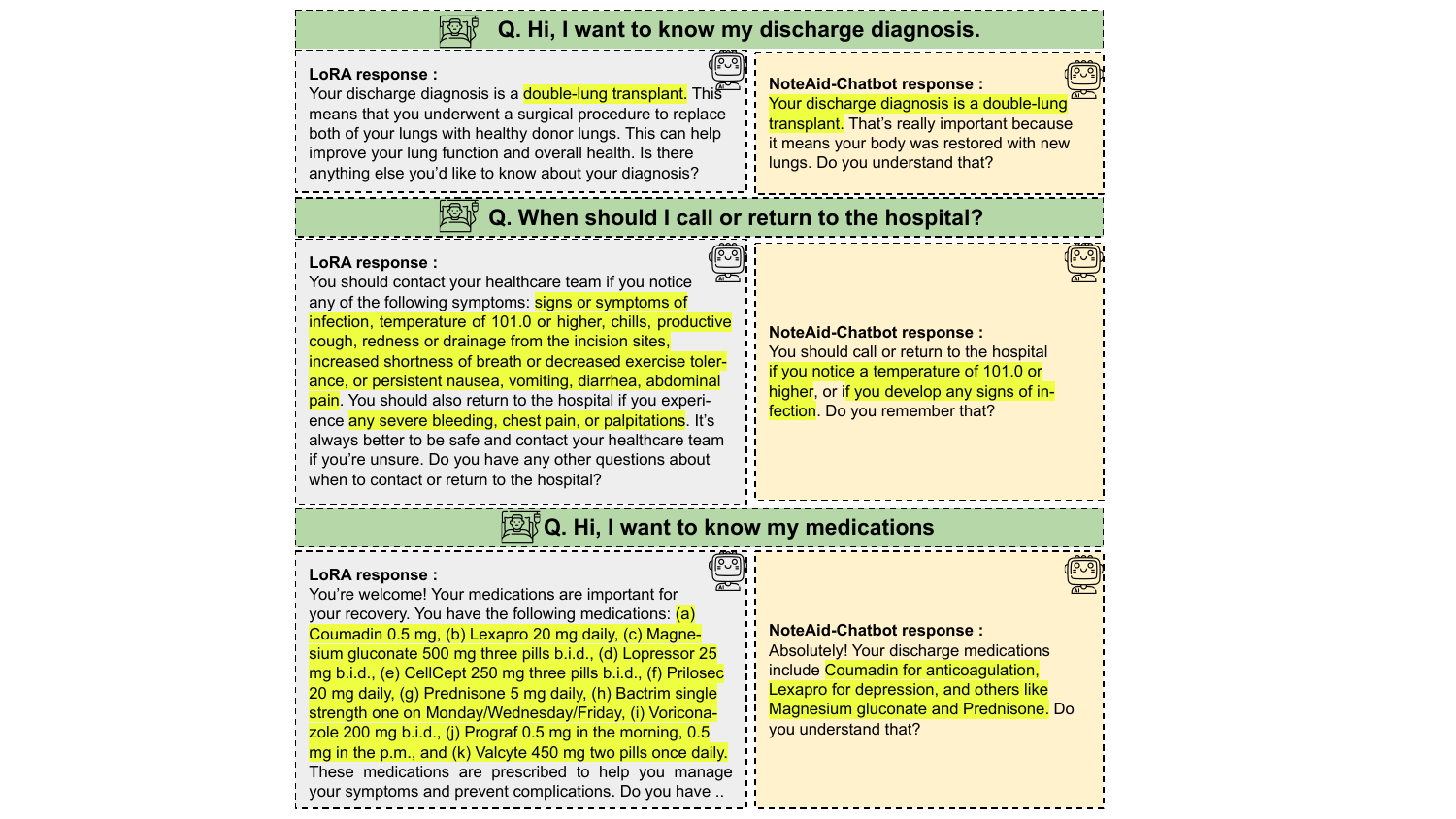}
    \caption{
    We compare the responses generated by NoteAid-Chatbot with those of the supervised fine-tuned LLaMA 3.2-3B-Instruct model. For each question posed, NoteAid-Chatbot consistently conveys equivalent content in a more concise and efficient manner.}
    \label{fig:response-comparison}
\end{figure}

As illustrated in Figure~\ref{fig:response-comparison}, NoteAid-Chatbot demonstrates the ability to generate more concise responses to identical questions, thereby enhancing textual readability. The examples further show that the chatbot delivers core information in a simplified manner, effectively addressing the essential components of each query.

In contrast, the supervised fine-tuned LLaMA 3.2-3B-Instruct model tends to include a greater volume of information from the discharge notes. While this may reflect comprehensive content coverage, presenting extensive information in a single turn is suboptimal for real-world patient communication. Given that typical patients possess a readability level corresponding to grades 6 to 8, as noted by \citet{okuhara_readability_2025}, it is more appropriate to communicate using simple, accessible language and to convey additional details incrementally across multiple conversational turns.

\subsection{Turing Test Results}%Human Interaction}

\begin{table}[h]
  \caption{Health literacy test and Comprehension test results in Turing test experiment. The whole group showed the same level of health literacy. Group B, showed a higher score than non-expert educators proving the capability of utilizing the chatbot for education scenarios. Group A: non-expert to human; Group B: NoteAid-Chatbot to human; Group C: expert to human}
  \label{tab-human-eval-results}
  \centering
  \begin{adjustbox}{width=\columnwidth, center}
  \begin{tabular}{cccc}
    % \cmidrule(r){1-2}
    \toprule
    \textbf{Group} & \textbf{n}  & \textbf{Health literacy} & \textbf{Comprehension score}  \\
    \midrule
    A & 5 &  33.200 & 0.650 \\
    B & 4 &  34.250 & 0.719  \\
    C & 4 &  35.750 & 0.750  \\
    \bottomrule
  \end{tabular}
  \end{adjustbox}
\end{table}

In the Turing test experiment, 13 students participated as the patient role, 5 students as a non-expert educator and 1 expert educator. Group C, representing expert-to-patient interactions, achieved the highest comprehension score of 0.750 (Table~\ref{tab-human-eval-results}). Group B, consisting of chatbot-to-patient interactions, attained a score of 0.719—achieving higher score than Group A (0.650), non-expert-to-patient interactions.\footnote{We could not conduct ANOVA test due to the small sample size.} While minor variations in health literacy scores were observed across groups, all participants demonstrated a comparable baseline level of health literacy based on the interpretation of S-TOFLA. 

Although the chatbot did not achieve the highest comprehension score, it demonstrated effectiveness in conveying discharge information. Notably, human educators were granted access to the discharge note prior to the interaction, whereas the chatbot engaged patients without prior exposure to the content. Given the model’s relatively small size ($\simeq3$B parameters), its performance approaches that of human educators. We hypothesize that incorporating test-time scaling techniques such as those proposed in~\cite{deepseek-ai_deepseek-r1_2025, muennighoff_s1_2025} could further enhance model performance. As these papers also suggest using simple methods to enhance the model's reasoning capabilities.

% \begin{table}
%   \caption{Usability survey results. Group A are the non-expert to human interaction group, Group B are chatbot to human interaction group and Group C are the non-expert to human interaction group.}
%   % \begin{adjustbox}{\linewidth}
%   \label{tab-human-eval-usability}
%   \centering
%   \begin{tabular}{p{0.6\linewidth} p{0.1\linewidth} p{0.1\linewidth} p{0.1\linewidth}}
%     % \cmidrule(r){1-2}
%     \toprule
%     \textbf{Questions} & \textbf{Group A} & \textbf{Group B} & \textbf{Group C} \\
%     \midrule
%     1. Interacting with the educator felt smooth. & 3.5  & 3 & 4 \\
%     2. The educator's conversation was consistent throughout the interaction & 3.25  & 2.75 & 4.667 \\
%     3. The educator responded in a way that felt natural and human-like. & 4 & 2.5 & 4.333 \\
%     4. I trusted the information provided by the educator. &  4.25 & 2.75 & 4  \\
%     5. The educator was able to understand and address my concerns. & 4.5 & 3 & 5  \\
%     6. The educator helped me better understand my discharge instructions. & 4.75 &  3.25 & 4.333 \\
%     7. I felt satisfied with the overall experience with the educator. & 4.5 & 3 & 3.333 \\
%     \midrule
%     Average & 4.107 & 2.892  & 4.380 \\
%     \bottomrule
%   \end{tabular}
%   % \end{adjustbox}
% \end{table}

\begin{table}
  \caption{Turing test result. Our participants could easily identify humans and the NoteAid-Chatbot. Group A : non-expert to human; Group B : NoteAid-Chatbot to human; Group C : expert to human}
  % \begin{adjustbox}{\linewidth}
 \begin{adjustbox}{width=0.95\columnwidth, center} 
  \label{tab:human-usability-q8}
  \centering
  \begin{tabular}{p{0.9
  \linewidth}p{0.03\linewidth}p{0.03\linewidth}p{0.03\linewidth}}
    % \cmidrule(r){1-2}
    \toprule
    \textbf{Overall did the educator feel like human?} & \textbf{A} & \textbf{B} & \textbf{C} \\
    \midrule
     Yes & 4  & 1 & 3 \\
     No & 1  & 2 & 0 \\
     Not Sure & 0  & 1 & 1 \\
    \bottomrule
  \end{tabular}
  % \end{adjustbox}
\end{adjustbox}
\end{table}
% We additionally conducted a survey based on the criteria outlined in Table~\ref{tab-human-eval-survey-contents}, with the results presented in Table~\ref{tab-human-eval-usability} and Table~\ref{tab-human-usability-q8}. Group C achieved the highest average usability score of 4.380, while the chatbot received a lower score of 2.892. Across all items in Table~\ref{tab-human-eval-usability}, the chatbot consistently received lower ratings in usability compared to the human-led groups. 

As illustrated in Table~\ref{tab:human-usability-q8}, the key limitation identified was the chatbot's lack of perceived “humanness”. In Group A and C, most of the participants could tell that their educator was a human. However in Group B, half of the students could tell that their educator was a chatbot. One potential explanation is the greater conversational flexibility observed in human interactions. During the experiments, patients often posed multiple questions or made compound utterances within a single turn. Human educators were able to respond dynamically and address each concern sequentially. In contrast, our Chatbot—trained on a strict multi-turn dialogue structure—was unable to replicate such adaptive conversational behavior.

\section{Related Work}

\noindent\textbf{Large Language Models in Healthcare:} 
LLMs such as GPT-4 and ChatGPT have shown promising performance in various healthcare tasks, particularly in answering medical questions with high accuracy and readability~\cite{achiam2023gpt,kung2023performance,goodman2023accuracy,decker2023large,ayers2023comparing,thirunavukarasu2023trialling,yao2025survey}. 
Domain-specific adaptations like Google’s \textit{Med-PaLM}~\cite{tu2024towards} have further improved safety and factual consistency by aligning LLMs with curated medical knowledge.
For instance, \textit{Med-PaLM} significantly reduced hallucinated or harmful responses and was rated more helpful by clinical professionals~\cite{singhal2023large}. 
However, general-purpose LLMs still suffer from issues~\cite{yang2025unveiling,jin2024hidden} such as factual errors and lack of personalization, making them unsuitable for patient-facing education tasks without proper alignment~\cite{sun2024effectiveness,aydin2024large}.

\noindent\textbf{Patient Education Chatbots and EHR Comprehension Tools:} 
Improving patient understanding of their EHRs has long been a goal of clinical NLP~\cite{nutbeam2023artificial,aydin2024large}. 
Early systems such as \textit{NoteAid}~\cite{polepalli2013improving} helped patients comprehend EHRs by linking medical jargon~\cite{kwon2022medjex} to lay definitions~\cite{yao2023readme}. 
More recently, \textit{PaniniQA}~\cite{cai2023paniniqa} introduced an interactive QA system that automatically generates patient-specific questions from discharge notes and verifies answers to reinforce understanding. 
Their method drew inspiration from dialogic reading~\cite{Whitehurst2002dialogic} and focused on guiding patients to uncover relationships between medical events via causal or correlational reasoning~\cite{cai2023paniniqa,lehman2022learning}.
Our work shares PaniniQA’s goal of enhancing post-visit comprehension through interactive conversation. However, there are three key differences. 
First, \textit{PaniniQA} relies on structured event and relation annotations to control question generation; in contrast, \textit{NoteAid-Chatbot} is trained end-to-end using synthetic conversations and reinforcement learning, requiring no expert-annotated supervision during training.
Second, while PaniniQA emphasizes question generation and selection, we frame the problem as a full multi-turn education task, where the chatbot dynamically guides patients through dialogue and reinforces learning based on test outcomes.
Third, our reward signals are verifiable and outcome-based, derived from simulated comprehension tests, enabling scalable RL-based alignment.

\noindent\textbf{Conversational Learning and Medical QA:} 
Recent research highlights the importance of learning through conversation, especially for patient education~\cite{golinkoff2019language,zhang2019dialogpt,xu2022fantastic}. \textit{Dialogic Reading}~\cite{Whitehurst2002dialogic,mol2008added,lever2011discussing} shows that guided dialogue can significantly improve knowledge retention in learners. While it is not always feasible to engage human clinicians in repeated one-on-one education sessions, chatbot-based dialogue systems can offer scalable and personalized alternatives~\cite{yao2021ai,cai2022generation}.
Unlike most clinical QA systems, which focus on fact retrieval or physician-style queries~\cite{pampari2018emrqa,jin2019pubmedqa,raghavan2021emrkbqa,yao2024medqa}, our system prioritizes education-oriented dialogue. This includes conversational strategies such as simplification, clarification, and empathy, which are crucial for improving patient comprehension~\cite{king_best_2013}. 
Moreover, while prior work has used LLMs for general and medical question generation~\cite{guo2024survey,klang2023advantages,yao2024mcqg}, these approaches rarely incorporate verifiable outcomes or educational objectives in their evaluation.

\section{Conclusion}

We present a multi-agent framework for automating the development of domain-specific, lightweight chatbot for patient education using RL-based alignment. The proposed approach utilizes supervised fine-tuning with synthetically generated data, followed by alignment through simple PPO technique. Our findings demonstrate that reinforcement learning significantly enhances the overall performance of the chatbot. Also, in a Turing test evaluation, NoteAid-Chatbot exhibited performance comparable to that of human educators.
% We evaluated the model across multiple criteria to assess its robustness and generalizability.

% First, the usability of the chatbot is still limited in a strict multi-turn of conversation. However, natural human  conversations are much dynamic where speakers talk more than one time in their utterances. To facilitate more natural and engaging dialogue, future work should focus on enabling and evaluating dynamic multi-turn conversational capabilities. 

\section{Ethical Concerns and Limitations}

\subsection{Ethical Concerns}
We have not yet implemented quantitative methods to detect or prevent hallucinated outputs—an inherent risk in deploying LLMs in clinical applications. In high-stakes environments, hallucinations could pose serious threats to patient safety. To mitigate this, we limited our implementation to discharge scenarios, where the contents of the note are known and can be verified. Recent work \cite{kim_medical_2025} has proposed using factuality metrics such as FactScore~\cite{min_factscore_2023} to assess and minimize hallucinations. Future iterations of this system should incorporate such mechanisms to ensure factual integrity.

Although we used the LLM-as-a-Judge framework to evaluate output quality, this method is not without limitations. Studies (e.g.,~\cite{lan_criticeval_2024}) have shown that evaluation outcomes may vary depending on the order of response presentation, introducing positional bias. While human evaluation could offer a more reliable benchmark [4], it is expensive and undermines the scalability of automated assessments. Pairwise evaluation strategies, as recommended by~\cite{ye_justice_2024}, may help reduce this bias while maintaining evaluation efficiency.

Fully autonomous use of NoteAid-Chat without human oversight may compromise patient safety. Clinical judgment and contextual understanding—especially in high-risk settings—remain beyond the capabilities of current AI systems. We therefore advocate for restricting the use of NoteAid-Chat to low-risk, informational contexts and avoiding decision-making on behalf of patients. Integrating human-in-the-loop oversight and applying mitigation strategies discussed above can make the system more robust and clinically responsible, serving as an assistive tool for both patients and healthcare providers.

\subsection{Limitations}

This study has several limitations. First, we did not explore alternative reinforcement learning (RL) alignment methods or incorporate recent advances in test-time optimization techniques. Investigating and comparing these approaches remains an important direction for future work. Second, during the reinforcement learning phase and subsequent simulations on $\mathcal{N_G}$ and $\mathcal{N_S}$, conversations were constrained to a maximum of 20 turns. Future iterations should enable the chatbot to autonomously determine appropriate termination points based on the conversational context. Third, the patient agent used during both training and simulation was implemented using GPT-4o-mini, roleplaying as a patient. However, its behavior may not accurately reflect real-world patient interactions. Future research will incorporate more robust and validated roleplay methodologies to create a more realistic simulation environment, thereby enhancing model performance. Finally, the human evaluation component was limited by a small sample size, with only five student participants per group. This narrow cohort does not capture the diversity of real-world patient populations. To improve the generalizability and validity of the findings, future studies will involve a larger and more representative sample.

\section*{Acknowledgments}

This work is supported by the Department of Veterans Affairs (VA), Veterans Health Administration, Office of Research and Development, VA Health Systems Research (IIR 24-083; I01HX003969). The views expressed in this article are those of the authors and do not necessarily reflect the position or policy of the Department of Veterans Affairs or the United States government.

% \section{References}
% \bibliographystyle{abbrvnat}
\bibliography{main, custom}

\begin{thebibliography}{85}
\providecommand{\natexlab}[1]{#1}

\bibitem[{Achiam et~al.(2023)Achiam, Adler, Agarwal, Ahmad, Akkaya, Aleman, Almeida, Altenschmidt, Altman, Anadkat et~al.}]{achiam2023gpt}
Josh Achiam, Steven Adler, Sandhini Agarwal, Lama Ahmad, Ilge Akkaya, Florencia~Leoni Aleman, Diogo Almeida, Janko Altenschmidt, Sam Altman, Shyamal Anadkat, and 1 others. 2023.
\newblock Gpt-4 technical report.
\newblock \emph{arXiv preprint arXiv:2303.08774}.

\bibitem[{Apter et~al.(2019)Apter, Localio, Morales, Han, Perez, Mullen, Rogers, Klusaritz, Howell, Canales, and Bryant-Stephens}]{apter_home_2019}
Andrea~J. Apter, A.~Russell Localio, Knashawn~H. Morales, Xiaoyan Han, Luzmercy Perez, Alyssa~N. Mullen, Marisa Rogers, Heather Klusaritz, John~T. Howell, Maryori~N. Canales, and Tyra Bryant-Stephens. 2019.
\newblock \href {https://doi.org/10.1016/j.jaci.2019.05.030} {Home visits for uncontrolled asthma among low-income adults with patient portal access}.
\newblock \emph{The Journal of Allergy and Clinical Immunology}, 144(3):846--853.e11.

\bibitem[{Assiri(2022)}]{assiri_impact_2022}
Ghadah Assiri. 2022.
\newblock \href {https://doi.org/10.1016/j.jsps.2022.01.001} {The {Impact} of {Patient} {Access} to {Their} {Electronic} {Health} {Record} on {Medication} {Management} {Safety}: {A} {Narrative} {Review}}.
\newblock \emph{Saudi pharmaceutical journal: SPJ: the official publication of the Saudi Pharmaceutical Society}, 30(3):185--194.

\bibitem[{Aydin et~al.(2024)Aydin, Karabacak, Vlachos, and Margetis}]{aydin2024large}
Serhat Aydin, Mert Karabacak, Victoria Vlachos, and Konstantinos Margetis. 2024.
\newblock Large language models in patient education: a scoping review of applications in medicine.
\newblock \emph{Frontiers in Medicine}, 11:1477898.

\bibitem[{Ayers et~al.(2023)Ayers, Poliak, Dredze, Leas, Zhu, Kelley, Faix, Goodman, Longhurst, Hogarth et~al.}]{ayers2023comparing}
John~W Ayers, Adam Poliak, Mark Dredze, Eric~C Leas, Zechariah Zhu, Jessica~B Kelley, Dennis~J Faix, Aaron~M Goodman, Christopher~A Longhurst, Michael Hogarth, and 1 others. 2023.
\newblock Comparing physician and artificial intelligence chatbot responses to patient questions posted to a public social media forum.
\newblock \emph{JAMA internal medicine}, 183(6):589--596.

\bibitem[{Baldry et~al.(1986)Baldry, Cheal, Fisher, Gillett, and Huet}]{baldry1986giving}
Molly Baldry, Carol Cheal, Brian Fisher, Myra Gillett, and Val Huet. 1986.
\newblock Giving patients their own records in general practice: experience of patients and staff.
\newblock \emph{British medical journal (Clinical research ed.)}, 292(6520):596--598.
\newblock Publisher: British Medical Journal Publishing Group.

\bibitem[{Bronson and O'Meara(1986)}]{bronson_impact_1986}
D.~L. Bronson and K.~O'Meara. 1986.
\newblock \href {https://doi.org/10.1007/BF02596322} {The impact of shared medical records on smoking awareness and behavior in ambulatory care}.
\newblock \emph{Journal of General Internal Medicine}, 1(1):34--37.

\bibitem[{Cai et~al.(2022)Cai, Liu, Bajracharya, Sills, Kapoor, Liu, Berlowitz, Levy, Pradhan, and Yu}]{cai2022generation}
Pengshan Cai, Fei Liu, Adarsha Bajracharya, Joe Sills, Alok Kapoor, Weisong Liu, Dan Berlowitz, David Levy, Richeek Pradhan, and Hong Yu. 2022.
\newblock Generation of patient after-visit summaries to support physicians.
\newblock In \emph{Proceedings of the 29th International Conference on Computational Linguistics (COLING)}.

\bibitem[{Cai et~al.(2023)Cai, Yao, Liu, Wang, Reilly, Zhou, Li, Cao, Kapoor, Bajracharya et~al.}]{cai2023paniniqa}
Pengshan Cai, Zonghai Yao, Fei Liu, Dakuo Wang, Meghan Reilly, Huixue Zhou, Lingxi Li, Yi~Cao, Alok Kapoor, Adarsha Bajracharya, and 1 others. 2023.
\newblock Paniniqa: Enhancing patient education through interactive question answering.
\newblock \emph{Transactions of the Association for Computational Linguistics}, 11:1518--1536.

\bibitem[{Chapman et~al.(2003)Chapman, Abraham, Jenkins, and Fallowfield}]{chapman2003lay}
Kristina Chapman, Charles Abraham, Valerie Jenkins, and Lesley Fallowfield. 2003.
\newblock Lay understanding of terms used in cancer consultations.
\newblock \emph{Psycho-Oncology: Journal of the Psychological, Social and Behavioral Dimensions of Cancer}, 12(6):557--566.
\newblock Publisher: Wiley Online Library.

\bibitem[{Decker et~al.(2023)Decker, Trang, Ramirez, Colley, Pierce, Coleman, Bongiovanni, Melton, and Wick}]{decker2023large}
Hannah Decker, Karen Trang, Joel Ramirez, Alexis Colley, Logan Pierce, Melissa Coleman, Tasce Bongiovanni, Genevieve~B Melton, and Elizabeth Wick. 2023.
\newblock Large language model- based chatbot vs surgeon-generated informed consent documentation for common procedures.
\newblock \emph{JAMA network open}, 6(10):e2336997--e2336997.

\bibitem[{DeepSeek-AI et~al.(2025)DeepSeek-AI, Guo, Yang, Zhang, Song, Zhang, Xu, Zhu, Ma, Wang, Bi, Zhang, Yu, Wu, Wu, Gou, Shao, Li, Gao, Liu, Xue, Wang, Wu, Feng, Lu, Zhao, Deng, Zhang, Ruan, Dai, Chen, Ji, Li, Lin, Dai, Luo, Hao, Chen, Li, Zhang, Bao, Xu, Wang, Ding, Xin, Gao, Qu, Li, Guo, Li, Wang, Chen, Yuan, Qiu, Li, Cai, Ni, Liang, Chen, Dong, Hu, Gao, Guan, Huang, Yu, Wang, Zhang, Zhao, Wang, Zhang, Xu, Xia, Zhang, Zhang, Tang, Li, Wang, Li, Tian, Huang, Zhang, Wang, Chen, Du, Ge, Zhang, Pan, Wang, Chen, Jin, Chen, Lu, Zhou, Chen, Ye, Wang, Yu, Zhou, Pan, Li, Zhou, Wu, Ye, Yun, Pei, Sun, Wang, Zeng, Zhao, Liu, Liang, Gao, Yu, Zhang, Xiao, An, Liu, Wang, Chen, Nie, Cheng, Liu, Xie, Liu, Yang, Li, Su, Lin, Li, Jin, Shen, Chen, Sun, Wang, Song, Zhou, Wang, Shan, Li, Wang, Wei, Zhang, Xu, Li, Zhao, Sun, Wang, Yu, Zhang, Shi, Xiong, He, Piao, Wang, Tan, Ma, Liu, Guo, Ou, Wang, Gong, Zou, He, Xiong, Luo, You, Liu, Zhou, Zhu, Xu, Huang, Li, Zheng, Zhu, Ma, Tang, Zha, Yan, Ren, Ren, Sha, Fu, Xu, Xie, Zhang,
  Hao, Ma, Yan, Wu, Gu, Zhu, Liu, Li, Xie, Song, Pan, Huang, Xu, Zhang, and Zhang}]{deepseek-ai_deepseek-r1_2025}
DeepSeek-AI, Daya Guo, Dejian Yang, Haowei Zhang, Junxiao Song, Ruoyu Zhang, Runxin Xu, Qihao Zhu, Shirong Ma, Peiyi Wang, Xiao Bi, Xiaokang Zhang, Xingkai Yu, Yu~Wu, Z.~F. Wu, Zhibin Gou, Zhihong Shao, Zhuoshu Li, Ziyi Gao, and 181 others. 2025.
\newblock \href {https://doi.org/10.48550/arXiv.2501.12948} {{DeepSeek}-{R1}: {Incentivizing} {Reasoning} {Capability} in {LLMs} via {Reinforcement} {Learning}}.
\newblock \emph{arXiv preprint}.
\newblock ArXiv:2501.12948 [cs].

\bibitem[{Delbanco et~al.(2012)Delbanco, Walker, Bell, Darer, Elmore, Farag, Feldman, Mejilla, Ngo, Ralston, Ross, Trivedi, Vodicka, and Leveille}]{delbanco_inviting_2012}
Tom Delbanco, Jan Walker, Sigall~K. Bell, Jonathan~D. Darer, Joann~G. Elmore, Nadine Farag, Henry~J. Feldman, Roanne Mejilla, Long Ngo, James~D. Ralston, Stephen~E. Ross, Neha Trivedi, Elisabeth Vodicka, and Suzanne~G. Leveille. 2012.
\newblock \href {https://doi.org/10.7326/0003-4819-157-7-201210020-00002} {Inviting patients to read their doctors' notes: a quasi-experimental study and a look ahead}.
\newblock \emph{Annals of Internal Medicine}, 157(7):461--470.

\bibitem[{DeSai et~al.(2021)DeSai, Janowiak, Secheli, Phelps, McDonald, Reed, and Blomkalns}]{desai_empowering_2021}
Charisma DeSai, Keri Janowiak, Beatrice Secheli, Eleanor Phelps, Sam McDonald, Gary Reed, and Andra Blomkalns. 2021.
\newblock \href {https://doi.org/10.1136/bmjoq-2021-001419} {Empowering patients: simplifying discharge instructions}.
\newblock \emph{BMJ Open Quality}, 10(3).
\newblock Publisher: British Medical Journal Publishing Group.

\bibitem[{Dewey(2004)}]{dewey_democracy_2004}
John Dewey. 2004.
\newblock \emph{Democracy and {Education}}.
\newblock Courier Corporation.
\newblock Google-Books-ID: 19ajcXf4MCYC.

\bibitem[{Dubey et~al.(2024)Dubey, Jauhri, Pandey, Kadian, Al-Dahle, Letman, Mathur, Schelten, Yang, Fan, and {others}}]{dubey2024llama}
Abhimanyu Dubey, Abhinav Jauhri, Abhinav Pandey, Abhishek Kadian, Ahmad Al-Dahle, Aiesha Letman, Akhil Mathur, Alan Schelten, Amy Yang, Angela Fan, and {others}. 2024.
\newblock The llama 3 herd of models.
\newblock \emph{arXiv preprint arXiv:2407.21783}.

\bibitem[{Elbourne et~al.(1987)Elbourne, Richardson, Chalmers, Waterhouse, and Holt}]{elbourne_newbury_1987}
D.~Elbourne, M.~Richardson, I.~Chalmers, I.~Waterhouse, and E.~Holt. 1987.
\newblock \href {https://doi.org/10.1111/j.1471-0528.1987.tb03165.x} {The {Newbury} {Maternity} {Care} {Study}: a randomized controlled trial to assess a policy of women holding their own obstetric records}.
\newblock \emph{British Journal of Obstetrics and Gynaecology}, 94(7):612--619.

\bibitem[{Flesch(2007)}]{flesch_flesch-kincaid_2007}
Rudolf Flesch. 2007.
\newblock \href {https://rockstar-english.com/lessons/advanced/12-Flesch_Kincaid_Readability_Test.pdf} {Flesch-{Kincaid} readability test}.
\newblock \emph{Retrieved October}, 26(3):2007.

\bibitem[{Golinkoff et~al.(2019)Golinkoff, Hoff, Rowe, Tamis-LeMonda, and Hirsh-Pasek}]{golinkoff2019language}
Roberta~Michnick Golinkoff, Erika Hoff, Meredith~L Rowe, Catherine~S Tamis-LeMonda, and Kathy Hirsh-Pasek. 2019.
\newblock Language matters: Denying the existence of the 30-million-word gap has serious consequences.
\newblock \emph{Child development}, 90(3):985--992.

\bibitem[{Goodman et~al.(2023)Goodman, Patrinely, Stone, Zimmerman, Donald, Chang, Berkowitz, Finn, Jahangir, Scoville et~al.}]{goodman2023accuracy}
Rachel~S Goodman, J~Randall Patrinely, Cosby~A Stone, Eli Zimmerman, Rebecca~R Donald, Sam~S Chang, Sean~T Berkowitz, Avni~P Finn, Eiman Jahangir, Elizabeth~A Scoville, and 1 others. 2023.
\newblock Accuracy and reliability of chatbot responses to physician questions.
\newblock \emph{JAMA network open}, 6(10):e2336483--e2336483.

\bibitem[{Griewing et~al.(2024)Griewing, Lechner, Gremke, Lukac, Janni, Wallwiener, Wagner, Hirsch, and Kuhn}]{griewing2024proof}
Sebastian Griewing, Fabian Lechner, Niklas Gremke, Stefan Lukac, Wolfgang Janni, Markus Wallwiener, Uwe Wagner, Martin Hirsch, and Sebastian Kuhn. 2024.
\newblock Proof-of-concept study of a small language model chatbot for breast cancer decision support--a transparent, source-controlled, explainable and data-secure approach.
\newblock \emph{Journal of Cancer Research and Clinical Oncology}, 150(10):451.

\bibitem[{Guo et~al.(2024)Guo, Liao, Li, and Chua}]{guo2024survey}
Shash Guo, Lizi Liao, Cuiping Li, and Tat-Seng Chua. 2024.
\newblock A survey on neural question generation: methods, applications, and prospects.
\newblock In \emph{Proceedings of the Thirty-Third International Joint Conference on Artificial Intelligence}, pages 8038--8047.

\bibitem[{Homer et~al.(1999)Homer, Davis, and Everitt}]{homer_introduction_1999}
Caroline~S.E. Homer, Gregory~K. Davis, and Louise~S. Everitt. 1999.
\newblock \href {https://doi.org/10.1111/j.1479-828X.1999.tb03445.x} {The {Introduction} of a {Woman}‐{Held} {Record} into a {Hospital} {Antenatal} {Clinic}: {The} {Bring} {Your} {Own} {Records} {Study}}.
\newblock \emph{Australian and New Zealand Journal of Obstetrics and Gynaecology}, 39(1):54--57.

\bibitem[{Hu et~al.(2021)Hu, Shen, Wallis, Allen-Zhu, Li, Wang, Wang, and Chen}]{hu_lora_2021}
Edward~J. Hu, Yelong Shen, Phillip Wallis, Zeyuan Allen-Zhu, Yuanzhi Li, Shean Wang, Lu~Wang, and Weizhu Chen. 2021.
\newblock \href {https://doi.org/10.48550/arXiv.2106.09685} {{LoRA}: {Low}-{Rank} {Adaptation} of {Large} {Language} {Models}}.
\newblock \emph{arXiv preprint}.
\newblock ArXiv:2106.09685 [cs].

\bibitem[{Hutson(2024)}]{hutson2024forget}
Matthew Hutson. 2024.
\newblock Forget chatgpt: why researchers now run small ais on their laptops.
\newblock \emph{Nature}, 633(8030):728--729.

\bibitem[{{Institute of Medicine (US) Committee on Health Literacy}(2004)}]{institute_of_medicine_us_committee_on_health_literacy_health_2004}
{Institute of Medicine (US) Committee on Health Literacy}. 2004.
\newblock \href {http://www.ncbi.nlm.nih.gov/books/NBK216032/} {\emph{Health {Literacy}: {A} {Prescription} to {End} {Confusion}}}.
\newblock National Academies Press (US), Washington (DC).

\bibitem[{Jin et~al.(2024)Jin, Chen, Zhou, Xu, Cheung, Chen, Summers, Rousseau, Ni, Landsman et~al.}]{jin2024hidden}
Qiao Jin, Fangyuan Chen, Yiliang Zhou, Ziyang Xu, Justin~M Cheung, Robert Chen, Ronald~M Summers, Justin~F Rousseau, Peiyun Ni, Marc~J Landsman, and 1 others. 2024.
\newblock Hidden flaws behind expert-level accuracy of multimodal gpt-4 vision in medicine.
\newblock \emph{npj Digital Medicine}, 7(1):190.

\bibitem[{Jin et~al.(2019)Jin, Dhingra, Liu, Cohen, and Lu}]{jin2019pubmedqa}
Qiao Jin, Bhuwan Dhingra, Zhengping Liu, William~W Cohen, and Xinghua Lu. 2019.
\newblock Pubmedqa: A dataset for biomedical research question answering.
\newblock \emph{arXiv preprint arXiv:1909.06146}.

\bibitem[{Johnson et~al.(2023)Johnson, Bulgarelli, Shen, Gayles, Shammout, Horng, Pollard, Hao, Moody, Gow, Lehman, Celi, and Mark}]{johnson_mimic-iv_2023}
Alistair E.~W. Johnson, Lucas Bulgarelli, Lu~Shen, Alvin Gayles, Ayad Shammout, Steven Horng, Tom~J. Pollard, Sicheng Hao, Benjamin Moody, Brian Gow, Li-wei~H. Lehman, Leo~A. Celi, and Roger~G. Mark. 2023.
\newblock \href {https://doi.org/10.1038/s41597-022-01899-x} {{MIMIC}-{IV}, a freely accessible electronic health record dataset}.
\newblock \emph{Scientific Data}, 10(1):1.
\newblock Number: 1 Publisher: Nature Publishing Group.

\bibitem[{Jones et~al.(1992)Jones, McGhee, and McGhee}]{jones1992patient}
RB~Jones, SM~McGhee, and D~McGhee. 1992.
\newblock Patient on-line access to medical records in general practice.
\newblock \emph{Health bulletin}, 50(2):143--150.

\bibitem[{Keselman et~al.(2007)Keselman, Slaughter, Arnott-Smith, Kim, Divita, Browne, Tsai, and Zeng-Treitler}]{keselman2007towards}
Alla Keselman, Laura Slaughter, Catherine Arnott-Smith, Hyeoneui Kim, Guy Divita, Allen Browne, Christopher Tsai, and Qing Zeng-Treitler. 2007.
\newblock Towards consumer-friendly {PHRs}: patients’ experience with reviewing their health records.
\newblock In \emph{{AMIA} annual symposium proceedings}, volume 2007, page 399. American Medical Informatics Association.

\bibitem[{Kim et~al.(2025)Kim, Jeong, Chen, Li, Lu, Alhamoud, Mun, Grau, Jung, Gameiro, Fan, Park, Lin, Yoon, Yoon, Sap, Tsvetkov, Liang, Xu, Liu, McDuff, Lee, Park, Tulebaev, and Breazeal}]{kim_medical_2025}
Yubin Kim, Hyewon Jeong, Shan Chen, Shuyue~Stella Li, Mingyu Lu, Kumail Alhamoud, Jimin Mun, Cristina Grau, Minseok Jung, Rodrigo Gameiro, Lizhou Fan, Eugene Park, Tristan Lin, Joonsik Yoon, Wonjin Yoon, Maarten Sap, Yulia Tsvetkov, Paul Liang, Xuhai Xu, and 6 others. 2025.
\newblock \href {https://doi.org/10.1101/2025.02.28.25323115} {Medical {Hallucination} in {Foundation} {Models} and {Their} {Impact} on {Healthcare}}.
\newblock Pages: 2025.02.28.25323115.

\bibitem[{King and Hoppe(2013)}]{king_best_2013}
Ann King and Ruth~B. Hoppe. 2013.
\newblock \href {https://doi.org/10.4300/JGME-D-13-00072.1} {“{Best} {Practice}” for {Patient}-{Centered} {Communication}: {A} {Narrative} {Review}}.
\newblock \emph{Journal of Graduate Medical Education}, 5(3):385--393.

\bibitem[{Klang et~al.(2023)Klang, Portugez, Gross, Brenner, Gilboa, Ortal, Ron, Robinzon, Meiri, Segal et~al.}]{klang2023advantages}
E~Klang, S~Portugez, R~Gross, A~Brenner, M~Gilboa, T~Ortal, S~Ron, V~Robinzon, H~Meiri, G~Segal, and 1 others. 2023.
\newblock Advantages and pitfalls in utilizing artificial intelligence for crafting medical examinations: a medical education pilot study with gpt-4.
\newblock \emph{BMC Medical Education}, 23.

\bibitem[{Kung et~al.(2023)Kung, Cheatham, Medenilla, Sillos, De~Leon, Elepa{\~n}o, Madriaga, Aggabao, Diaz-Candido, Maningo et~al.}]{kung2023performance}
Tiffany~H Kung, Morgan Cheatham, Arielle Medenilla, Czarina Sillos, Lorie De~Leon, Camille Elepa{\~n}o, Maria Madriaga, Rimel Aggabao, Giezel Diaz-Candido, James Maningo, and 1 others. 2023.
\newblock Performance of chatgpt on usmle: potential for ai-assisted medical education using large language models.
\newblock \emph{PLoS digital health}, 2(2):e0000198.

\bibitem[{Kutner et~al.(2006)Kutner, Greenburg, Jin, and Paulsen}]{kutner2006health}
Mark Kutner, Elizabeth Greenburg, Ying Jin, and Christine Paulsen. 2006.
\newblock The health literacy of america's adults: {Results} from the 2003 national assessment of adult literacy. {NCES} 2006-483.
\newblock \emph{National Center for education statistics}.
\newblock Publisher: ERIC.

\bibitem[{Kwon et~al.(2022)Kwon, Yao, Jordan, Levy, Corner, and Yu}]{kwon2022medjex}
Sunjae Kwon, Zonghai Yao, Harmon Jordan, David Levy, Brian Corner, and Hong Yu. 2022.
\newblock Medjex: A medical jargon extraction model with wiki’s hyperlink span and contextualized masked language model score.
\newblock In \emph{Proceedings of the 2022 Conference on Empirical Methods in Natural Language Processing}, pages 11733--11751.

\bibitem[{Labrak et~al.(2024)Labrak, Bazoge, Morin, Gourraud, Rouvier, and Dufour}]{labrak_biomistral_2024}
Yanis Labrak, Adrien Bazoge, Emmanuel Morin, Pierre-Antoine Gourraud, Mickael Rouvier, and Richard Dufour. 2024.
\newblock \href {https://doi.org/10.48550/arXiv.2402.10373} {{BioMistral}: {A} {Collection} of {Open}-{Source} {Pretrained} {Large} {Language} {Models} for {Medical} {Domains}}.
\newblock \emph{arXiv preprint}.
\newblock ArXiv:2402.10373 [cs].

\bibitem[{Lan et~al.(2024)Lan, Zhang, Xu, Huang, Lin, Chen, and Mao}]{lan_criticeval_2024}
Tian Lan, Wenwei Zhang, Chen Xu, Heyan Huang, Dahua Lin, Kai Chen, and Xian-ling Mao. 2024.
\newblock \href {https://doi.org/10.48550/arXiv.2402.13764} {{CriticEval}: {Evaluating} {Large} {Language} {Model} as {Critic}}.
\newblock \emph{arXiv preprint}.
\newblock ArXiv:2402.13764 [cs].

\bibitem[{Lehman et~al.(2022)Lehman, Lialin, Legaspi, Sy, Pile, Alberto, Ragasa, Puyat, Alberto, Alfonso et~al.}]{lehman2022learning}
Eric Lehman, Vladislav Lialin, Katelyn~Y Legaspi, Anne Janelle~R Sy, Patricia Therese~S Pile, Nicole Rose~I Alberto, Richard Raymund~R Ragasa, Corinna Victoria~M Puyat, Isabelle Rose~I Alberto, Pia Gabrielle~I Alfonso, and 1 others. 2022.
\newblock Learning to ask like a physician.
\newblock \emph{arXiv preprint arXiv:2206.02696}.

\bibitem[{Lerner et~al.(2000)Lerner, Jehle, Janicke, and Moscati}]{lerner2000medical}
E~Brooke Lerner, Dietrich~VK Jehle, David~M Janicke, and Ronald~M Moscati. 2000.
\newblock Medical communication: do our patients understand?
\newblock \emph{The American journal of emergency medicine}, 18(7):764--766.
\newblock Publisher: Elsevier.

\bibitem[{Lever and S{\'e}n{\'e}chal(2011)}]{lever2011discussing}
Rosemary Lever and Monique S{\'e}n{\'e}chal. 2011.
\newblock Discussing stories: On how a dialogic reading intervention improves kindergartners’ oral narrative construction.
\newblock \emph{Journal of experimental child psychology}, 108(1):1--24.

\bibitem[{Lin(2004)}]{lin_rouge_2004}
Chin-Yew Lin. 2004.
\newblock \href {https://aclanthology.org/W04-1013.pdf} {Rouge: {A} package for automatic evaluation of summaries}.
\newblock In \emph{Text summarization branches out}, pages 74--81.

\bibitem[{Lu et~al.(2024)Lu, Li, Cai, Yi, Liu, Zhang, Lane, and Xu}]{lu2024small}
Zhenyan Lu, Xiang Li, Dongqi Cai, Rongjie Yi, Fangming Liu, Xiwen Zhang, Nicholas~D Lane, and Mengwei Xu. 2024.
\newblock Small language models: Survey, measurements, and insights.
\newblock \emph{arXiv preprint arXiv:2409.15790}.

\bibitem[{Min et~al.(2023)Min, Krishna, Lyu, Lewis, Yih, Koh, Iyyer, Zettlemoyer, and Hajishirzi}]{min_factscore_2023}
Sewon Min, Kalpesh Krishna, Xinxi Lyu, Mike Lewis, Wen-tau Yih, Pang~Wei Koh, Mohit Iyyer, Luke Zettlemoyer, and Hannaneh Hajishirzi. 2023.
\newblock \href {https://doi.org/10.48550/arXiv.2305.14251} {{FActScore}: {Fine}-grained {Atomic} {Evaluation} of {Factual} {Precision} in {Long} {Form} {Text} {Generation}}.
\newblock \emph{arXiv preprint}.
\newblock ArXiv:2305.14251 [cs].

\bibitem[{Mol et~al.(2008)Mol, Bus, De~Jong, and Smeets}]{mol2008added}
Suzanne~E Mol, Adriana~G Bus, Maria~T De~Jong, and Daisy~JH Smeets. 2008.
\newblock Added value of dialogic parent--child book readings: A meta-analysis.
\newblock \emph{Early education and development}, 19(1):7--26.

\bibitem[{Muennighoff et~al.(2025)Muennighoff, Yang, Shi, Li, Fei-Fei, Hajishirzi, Zettlemoyer, Liang, Candès, and Hashimoto}]{muennighoff_s1_2025}
Niklas Muennighoff, Zitong Yang, Weijia Shi, Xiang~Lisa Li, Li~Fei-Fei, Hannaneh Hajishirzi, Luke Zettlemoyer, Percy Liang, Emmanuel Candès, and Tatsunori Hashimoto. 2025.
\newblock \href {https://doi.org/10.48550/arXiv.2501.19393} {s1: {Simple} test-time scaling}.
\newblock \emph{arXiv preprint}.
\newblock ArXiv:2501.19393 [cs].

\bibitem[{Nutbeam(2023)}]{nutbeam2023artificial}
Don Nutbeam. 2023.
\newblock Artificial intelligence and health literacy—proceed with caution.
\newblock \emph{Health Literacy and Communication Open}, 1(1):2263355.

\bibitem[{Okuhara et~al.(2025)Okuhara, Furukawa, Okada, Yokota, and Kiuchi}]{okuhara_readability_2025}
Tsuyoshi Okuhara, Emi Furukawa, Hiroko Okada, Rie Yokota, and Takahiro Kiuchi. 2025.
\newblock \href {https://doi.org/10.1016/j.pec.2025.108656} {Readability of written information for patients across 30 years: {A} systematic review of systematic reviews}.
\newblock \emph{Patient Education and Counseling}, 135:108656.

\bibitem[{{OpenAI}()}]{openai_website}
{OpenAI}.
\newblock \href {https://openai.com/} {{OpenAI} website}.

\bibitem[{Ouyang et~al.(2022)Ouyang, Wu, Jiang, Almeida, Wainwright, Mishkin, Zhang, Agarwal, Slama, Ray, Schulman, Hilton, Kelton, Miller, Simens, Askell, Welinder, Christiano, Leike, and Lowe}]{ouyang_training_2022}
Long Ouyang, Jeff Wu, Xu~Jiang, Diogo Almeida, Carroll~L. Wainwright, Pamela Mishkin, Chong Zhang, Sandhini Agarwal, Katarina Slama, Alex Ray, John Schulman, Jacob Hilton, Fraser Kelton, Luke Miller, Maddie Simens, Amanda Askell, Peter Welinder, Paul Christiano, Jan Leike, and Ryan Lowe. 2022.
\newblock \href {https://doi.org/10.48550/arXiv.2203.02155} {Training language models to follow instructions with human feedback}.
\newblock \emph{arXiv preprint}.
\newblock ArXiv:2203.02155 [cs].

\bibitem[{Pampari et~al.(2018)Pampari, Raghavan, Liang, and Peng}]{pampari2018emrqa}
Anusri Pampari, Preethi Raghavan, Jennifer Liang, and Jian Peng. 2018.
\newblock emrqa: A large corpus for question answering on electronic medical records.
\newblock \emph{arXiv preprint arXiv:1809.00732}.

\bibitem[{Papineni et~al.(2002)Papineni, Roukos, Ward, and Zhu}]{papineni_bleu_2002}
Kishore Papineni, Salim Roukos, Todd Ward, and Wei-Jing Zhu. 2002.
\newblock \href {https://aclanthology.org/P02-1040.Pdf} {Bleu: a method for automatic evaluation of machine translation}.
\newblock In \emph{Proceedings of the 40th annual meeting of the {Association} for {Computational} {Linguistics}}, pages 311--318.

\bibitem[{Parker et~al.(1995)Parker, Baker, Williams, and Nurss}]{parker_test_1995}
R.~M. Parker, D.~W. Baker, M.~V. Williams, and J.~R. Nurss. 1995.
\newblock \href {https://doi.org/10.1007/BF02640361} {The test of functional health literacy in adults: a new instrument for measuring patients' literacy skills}.
\newblock \emph{Journal of General Internal Medicine}, 10(10):537--541.

\bibitem[{Polepalli~Ramesh et~al.(2013)Polepalli~Ramesh, Houston, Brandt, Fang, and Yu}]{polepalli2013improving}
Balaji Polepalli~Ramesh, Thomas Houston, Cynthia Brandt, Hua Fang, and Hong Yu. 2013.
\newblock Improving patients' electronic health record comprehension with noteaid.
\newblock In \emph{MEDINFO 2013}, pages 714--718. IOS Press.

\bibitem[{Pyper et~al.(2004)Pyper, Amery, Watson, and Crook}]{pyper2004patients}
Cecilia Pyper, Justin Amery, Marion Watson, and Claire Crook. 2004.
\newblock Patients' experiences when accessing their on-line electronic patient records in primary care.
\newblock \emph{British Journal of General Practice}, 54(498):38--43.
\newblock Publisher: British Journal of General Practice.

\bibitem[{Raghavan et~al.(2021)Raghavan, Liang, Mahajan, Chandra, and Szolovits}]{raghavan2021emrkbqa}
Preethi Raghavan, Jennifer~J Liang, Diwakar Mahajan, Rachita Chandra, and Peter Szolovits. 2021.
\newblock emrkbqa: A clinical knowledge-base question answering dataset.
\newblock Association for Computational Linguistics.

\bibitem[{Sarkar et~al.(2010)Sarkar, Karter, Liu, Adler, Nguyen, Lopez, and Schillinger}]{sarkar2010literacy}
Urmimala Sarkar, Andrew~J Karter, Jennifer~Y Liu, Nancy~E Adler, Robert Nguyen, Andrea Lopez, and Dean Schillinger. 2010.
\newblock The literacy divide: health literacy and the use of an internet-based patient portal in an integrated health system—results from the {Diabetes} {Study} of {Northern} {California} ({DISTANCE}).
\newblock \emph{Journal of health communication}, 15(S2):183--196.
\newblock Publisher: Taylor \& Francis.

\bibitem[{Schulman et~al.(2017)Schulman, Wolski, Dhariwal, Radford, and Klimov}]{schulman_proximal_2017}
John Schulman, Filip Wolski, Prafulla Dhariwal, Alec Radford, and Oleg Klimov. 2017.
\newblock \href {https://doi.org/10.48550/arXiv.1707.06347} {Proximal {Policy} {Optimization} {Algorithms}}.
\newblock \emph{arXiv preprint}.
\newblock ArXiv:1707.06347 [cs].

\bibitem[{Sharples(2005)}]{sharples_learning_2005}
Mike Sharples. 2005.
\newblock \href {https://www.academia.edu/download/4272835/10.1.1.134.4715.pdf} {\emph{Learning as conversation transforming education in the mobile age}}.
\newblock na.

\bibitem[{Singhal et~al.(2023)Singhal, Azizi, Tu, Mahdavi, Wei, Chung, Scales, Tanwani, Cole-Lewis, Pfohl et~al.}]{singhal2023large}
Karan Singhal, Shekoofeh Azizi, Tao Tu, S~Sara Mahdavi, Jason Wei, Hyung~Won Chung, Nathan Scales, Ajay Tanwani, Heather Cole-Lewis, Stephen Pfohl, and 1 others. 2023.
\newblock Large language models encode clinical knowledge.
\newblock \emph{Nature}, 620(7972):172--180.

\bibitem[{Stossel et~al.(2012)Stossel, Segar, Gliatto, Fallar, and Karani}]{stossel_readability_2012}
Lauren~M. Stossel, Nora Segar, Peter Gliatto, Robert Fallar, and Reena Karani. 2012.
\newblock \href {https://doi.org/10.1007/s11606-012-2046-0} {Readability of {Patient} {Education} {Materials} {Available} at the {Point} of {Care}}.
\newblock \emph{Journal of General Internal Medicine}, 27(9):1165--1170.

\bibitem[{Sun et~al.(2024)Sun, Reiter, Kiltie, Ramsay, Duncan, Murchie, and Adam}]{sun2024effectiveness}
Mengxuan Sun, Ehud Reiter, Anne~E Kiltie, George Ramsay, Lisa Duncan, Peter Murchie, and Rosalind Adam. 2024.
\newblock Effectiveness of chatgpt in explaining complex medical reports to patients.
\newblock \emph{arXiv preprint arXiv:2406.15963}.

\bibitem[{Thirunavukarasu et~al.(2023)Thirunavukarasu, Hassan, Mahmood, Sanghera, Barzangi, El~Mukashfi, and Shah}]{thirunavukarasu2023trialling}
Arun~James Thirunavukarasu, Refaat Hassan, Shathar Mahmood, Rohan Sanghera, Kara Barzangi, Mohanned El~Mukashfi, and Sachin Shah. 2023.
\newblock Trialling a large language model (chatgpt) in general practice with the applied knowledge test: observational study demonstrating opportunities and limitations in primary care.
\newblock \emph{JMIR Medical Education}, 9(1):e46599.

\bibitem[{Tieu et~al.(2015)Tieu, Sarkar, Schillinger, Ralston, Ratanawongsa, Pasick, and Lyles}]{tieu2015barriers}
Lina Tieu, Urmimala Sarkar, Dean Schillinger, James~D Ralston, Neda Ratanawongsa, Rena Pasick, and Courtney~R Lyles. 2015.
\newblock Barriers and facilitators to online portal use among patients and caregivers in a safety net health care system: a qualitative study.
\newblock \emph{Journal of medical Internet research}, 17(12):e275.
\newblock Publisher: JMIR Publications Inc. Toronto, Canada.

\bibitem[{Tu et~al.(2024)Tu, Azizi, Driess, Schaekermann, Amin, Chang, Carroll, Lau, Tanno, Ktena et~al.}]{tu2024towards}
Tao Tu, Shekoofeh Azizi, Danny Driess, Mike Schaekermann, Mohamed Amin, Pi-Chuan Chang, Andrew Carroll, Charles Lau, Ryutaro Tanno, Ira Ktena, and 1 others. 2024.
\newblock Towards generalist biomedical ai.
\newblock \emph{Nejm Ai}, 1(3):AIoa2300138.

\bibitem[{Tu et~al.(2025)Tu, Schaekermann, Palepu, Saab, Freyberg, Tanno, Wang, Li, Amin, Cheng et~al.}]{tu_towards_2025}
Tao Tu, Mike Schaekermann, Anil Palepu, Khaled Saab, Jan Freyberg, Ryutaro Tanno, Amy Wang, Brenna Li, Mohamed Amin, Yong Cheng, and 1 others. 2025.
\newblock Towards conversational diagnostic artificial intelligence.
\newblock \emph{Nature}, pages 1--9.

\bibitem[{Wang et~al.(2024)Wang, Dang, Kostakos, and Jia}]{wang2024efficient}
Xin Wang, Ting Dang, Vassilis Kostakos, and Hong Jia. 2024.
\newblock Efficient and personalized mobile health event prediction via small language models.
\newblock In \emph{Proceedings of the 30th Annual International Conference on Mobile Computing and Networking}, pages 2353--2358.

\bibitem[{Whitehurst(2002)}]{Whitehurst2002dialogic}
Grover~J. Whitehurst. 2002.
\newblock Dialogic reading: {An} effective way to read aloud with young children.
\newblock \emph{https://www.readingrockets.org/article/dialogic-reading-effective-way-read-aloud-young-children}.

\bibitem[{Xu et~al.(2024)Xu, Fu, Gao, Ye, Liu, Mei, Wang, Yu, and Wu}]{xu_is_2024}
Shusheng Xu, Wei Fu, Jiaxuan Gao, Wenjie Ye, Weilin Liu, Zhiyu Mei, Guangju Wang, Chao Yu, and Yi~Wu. 2024.
\newblock \href {https://doi.org/10.48550/arXiv.2404.10719} {Is {DPO} {Superior} to {PPO} for {LLM} {Alignment}? {A} {Comprehensive} {Study}}.
\newblock \emph{arXiv preprint}.
\newblock ArXiv:2404.10719 [cs].

\bibitem[{Xu et~al.(2022)Xu, Wang, Yu, Ritchie, Yao, Wu, Zhang, Li, Bradford, Sun et~al.}]{xu2022fantastic}
Ying Xu, Dakuo Wang, Mo~Yu, Daniel Ritchie, Bingsheng Yao, Tongshuang Wu, Zheng Zhang, Toby Jia-Jun Li, Nora Bradford, Branda Sun, and 1 others. 2022.
\newblock Fantastic questions and where to find them: Fairytaleqa--an authentic dataset for narrative comprehension.
\newblock \emph{arXiv preprint arXiv:2203.13947}.

\bibitem[{Yang et~al.(2025{\natexlab{a}})Yang, Li, Yang, Zhang, Hui, Zheng, Yu, Gao, Huang, Lv, Zheng, Liu, Zhou, Huang, Hu, Ge, Wei, Lin, Tang, Yang, Tu, Zhang, Yang, Yang, Zhou, Zhou, Lin, Dang, Bao, Yang, Yu, Deng, Li, Xue, Li, Zhang, Wang, Zhu, Men, Gao, Liu, Luo, Li, Tang, Yin, Ren, Wang, Zhang, Ren, Fan, Su, Zhang, Zhang, Wan, Liu, Wang, Cui, Zhang, Zhou, and Qiu}]{yang_qwen3_2025}
An~Yang, Anfeng Li, Baosong Yang, Beichen Zhang, Binyuan Hui, Bo~Zheng, Bowen Yu, Chang Gao, Chengen Huang, Chenxu Lv, Chujie Zheng, Dayiheng Liu, Fan Zhou, Fei Huang, Feng Hu, Hao Ge, Haoran Wei, Huan Lin, Jialong Tang, and 41 others. 2025{\natexlab{a}}.
\newblock \href {https://doi.org/10.48550/arXiv.2505.09388} {Qwen3 {Technical} {Report}}.
\newblock \emph{arXiv preprint}.
\newblock ArXiv:2505.09388 [cs].

\bibitem[{Yang et~al.(2025{\natexlab{b}})Yang, Zhou, Qi, Zhen, Sun, Shi, Su, and Yang}]{yang_aligning_2025}
Lingrui Yang, Yuxing Zhou, Jun Qi, Xiantong Zhen, Li~Sun, Shan Shi, Qinghua Su, and Xuedong Yang. 2025{\natexlab{b}}.
\newblock \href {https://doi.org/10.1016/j.ejrad.2025.111984} {Aligning large language models with radiologists by reinforcement learning from {AI} feedback for chest {CT} reports}.
\newblock \emph{European Journal of Radiology}, 184:111984.

\bibitem[{Yang et~al.(2025{\natexlab{c}})Yang, Yao, Tasmin, Vashisht, Jang, Ouyang, Wang, McManus, Berlowitz, and Yu}]{yang2025unveiling}
Zhichao Yang, Zonghai Yao, Mahbuba Tasmin, Parth Vashisht, Won~Seok Jang, Feiyun Ouyang, Beining Wang, David McManus, Dan Berlowitz, and Hong Yu. 2025{\natexlab{c}}.
\newblock Unveiling gpt-4v's hidden challenges behind high accuracy on usmle questions: Observational study.
\newblock \emph{Journal of Medical Internet Research}, 27:e65146.

\bibitem[{Yao et~al.(2021)Yao, Wang, Wu, Zhang, Li, Yu, and Xu}]{yao2021ai}
Bingsheng Yao, Dakuo Wang, Tongshuang Wu, Zheng Zhang, Toby Jia-Jun Li, Mo~Yu, and Ying Xu. 2021.
\newblock It is ai's turn to ask humans a question: question-answer pair generation for children's story books.
\newblock \emph{arXiv preprint arXiv:2109.03423}.

\bibitem[{Yao et~al.(2024{\natexlab{a}})Yao, Kantu, Wei, Tran, Duan, Kwon, Yang, and Yu}]{yao2023readme}
Zonghai Yao, Nandyala~Siddharth Kantu, Guanghao Wei, Hieu Tran, Zhangqi Duan, Sunjae Kwon, Zhichao Yang, and Hong Yu. 2024{\natexlab{a}}.
\newblock Readme: Bridging medical jargon and lay understanding for patient education through data-centric nlp.
\newblock In \emph{Findings of the Association for Computational Linguistics: EMNLP 2024}, pages 12609--12629.

\bibitem[{Yao et~al.(2025)Yao, Parashar, Zhou, Jang, Ouyang, Yang, and Yu}]{yao2024mcqg}
Zonghai Yao, Aditya Parashar, Huixue Zhou, Won~Seok Jang, Feiyun Ouyang, Zhichao Yang, and Hong Yu. 2025.
\newblock Mcqg-srefine: Multiple choice question generation and evaluation with iterative self-critique, correction, and comparison feedback.
\newblock In \emph{Proceedings of the 2025 Conference of the Nations of the Americas Chapter of the Association for Computational Linguistics: Human Language Technologies (Volume 1: Long Papers)}, pages 10728--10777.

\bibitem[{Yao and Yu(2025)}]{yao2025survey}
Zonghai Yao and Hong Yu. 2025.
\newblock A survey on llm-based multi-agent ai hospital.

\bibitem[{Yao et~al.(2024{\natexlab{b}})Yao, Zhang, Tang, Bian, Zhao, Yang, Wang, Zhou, Jang, Ouyang et~al.}]{yao2024medqa}
Zonghai Yao, Zihao Zhang, Chaolong Tang, Xingyu Bian, Youxia Zhao, Zhichao Yang, Junda Wang, Huixue Zhou, Won~Seok Jang, Feiyun Ouyang, and 1 others. 2024{\natexlab{b}}.
\newblock Medqa-cs: Benchmarking large language models clinical skills using an ai-sce framework.
\newblock \emph{arXiv preprint arXiv:2410.01553}.

\bibitem[{Ye et~al.(2024)Ye, Wang, Huang, Chen, Zhang, Moniz, Gao, Geyer, Huang, Chen, Chawla, and Zhang}]{ye_justice_2024}
Jiayi Ye, Yanbo Wang, Yue Huang, Dongping Chen, Qihui Zhang, Nuno Moniz, Tian Gao, Werner Geyer, Chao Huang, Pin-Yu Chen, Nitesh~V. Chawla, and Xiangliang Zhang. 2024.
\newblock \href {https://doi.org/10.48550/arXiv.2410.02736} {Justice or {Prejudice}? {Quantifying} {Biases} in {LLM}-as-a-{Judge}}.
\newblock \emph{arXiv preprint}.
\newblock ArXiv:2410.02736 [cs].

\bibitem[{Zarcadoolas et~al.(2013)Zarcadoolas, Vaughon, Czaja, Levy, and Rockoff}]{zarcadoolas2013consumers}
Christina Zarcadoolas, Wendy~L Vaughon, Sara~J Czaja, Joslyn Levy, and Maxine~L Rockoff. 2013.
\newblock Consumers' perceptions of patient-accessible electronic medical records.
\newblock \emph{Journal of medical Internet research}, 15(8):e168.
\newblock Publisher: JMIR Publications Inc. Toronto, Canada.

\bibitem[{Zhang et~al.(2025)Zhang, Liu, Qin, Naumann, and Poon}]{zhang_med-rlvr_2025}
Sheng Zhang, Qianchu Liu, Guanghui Qin, Tristan Naumann, and Hoifung Poon. 2025.
\newblock \href {https://doi.org/10.48550/arXiv.2502.19655} {Med-{RLVR}: {Emerging} {Medical} {Reasoning} from a {3B} base model via reinforcement {Learning}}.
\newblock \emph{arXiv preprint}.
\newblock ArXiv:2502.19655 [cs].

\bibitem[{Zhang et~al.(2020)Zhang, Kishore, Wu, Weinberger, and Artzi}]{zhang_bertscore_2020}
Tianyi Zhang, Varsha Kishore, Felix Wu, Kilian~Q. Weinberger, and Yoav Artzi. 2020.
\newblock \href {https://doi.org/10.48550/arXiv.1904.09675} {{BERTScore}: {Evaluating} {Text} {Generation} with {BERT}}.
\newblock \emph{arXiv preprint}.
\newblock ArXiv:1904.09675 [cs].

\bibitem[{Zhang et~al.(2019)Zhang, Sun, Galley, Chen, Brockett, Gao, Gao, Liu, and Dolan}]{zhang2019dialogpt}
Yizhe Zhang, Siqi Sun, Michel Galley, Yen-Chun Chen, Chris Brockett, Xiang Gao, Jianfeng Gao, Jingjing Liu, and Bill Dolan. 2019.
\newblock Dialogpt: Large-scale generative pre-training for conversational response generation.
\newblock \emph{arXiv preprint arXiv:1911.00536}.

\bibitem[{Zheng et~al.(2023)Zheng, Chiang, Sheng, Zhuang, Wu, Zhuang, Lin, Li, Li, Xing, Zhang, Gonzalez, and Stoica}]{zheng_judging_2023}
Lianmin Zheng, Wei-Lin Chiang, Ying Sheng, Siyuan Zhuang, Zhanghao Wu, Yonghao Zhuang, Zi~Lin, Zhuohan Li, Dacheng Li, Eric~P. Xing, Hao Zhang, Joseph~E. Gonzalez, and Ion Stoica. 2023.
\newblock \href {https://doi.org/10.48550/arXiv.2306.05685} {Judging {LLM}-as-a-{Judge} with {MT}-{Bench} and {Chatbot} {Arena}}.
\newblock \emph{arXiv preprint}.
\newblock ArXiv:2306.05685 [cs].

\end{thebibliography}

\appendix
\label{appendix}

\appendix
\section{Dataset Evaluation}
\label{appendix:dataset_evaluation}

\subsection{Dataset Quration}

For ${Comp}_{G}$ and ${Comp}_{S}$, we ensure the quality with different measures. 

% $$\mathcal{D}_{Gold} = \{(\mathcal{N}^i_{G}, Q^i) | i \in \text{[1,100]} \}$$

We take two steps to generate and evaluate the quality of $\mathcal{D}_{Gold}$. First by asking students to annotate 5-10 questionnaires. And then, 3 experts will go through the generated datasets and evaluate and comment or modify the questionnaires. The students were all PhD students majoring in Computer Science in the United States. The expert annotators were 2 nurse professors in the United States and 1 doctor from South Korea.

For each $\mathcal{N}_{G}$, we instructed the annotators to generate 5 to 10 multiple choice questions with 3 choices; answer, distractor and irrelevant as shown in figure~\ref{fig:guideline-generation-for-Q} We asked three medical experts to go through the questions and the questionnaires to validate the quality of the annotated $Q \in {Comp}_{G}$. We asked them to modify or leave comments to $Q$ and made the changes according to their comments (Figure~\ref{fig:silver-expert-annotation-guideline}).

For the generated dataset, ${Comp}_{S}$, we first ensured that we have diverse $\mathcal{N}_S$ generated by GPT-4o-mini. We use the prompt in Figure~\ref{fig:synthetic-note-generation-prompt} to generate the synthetic notes.
To do so, we prepared specific demographic criterias-Age, Gender, Ethnicity, Disease category, Chief Complaints, Associated Procedures. For Disease category, Chief Complaints and Associated Procedures, we kept combinations that were clinically plausible since some combinations could irrelevant in clinical perspective (Table \ref{tab:silver-note-combinations}). And by mixing the combinations of these criterias with a predefined distribution shown in Table~\ref{tab:silver-note-demographic} for each category, we instruct GPT-4o-mini to generate a discharge note that contains the six medical content categories suggested in Table~\ref{tab1-medcontent-strategy}. We follow the demographic distribution of MIMIC-IV dataset~\cite{johnson_mimic-iv_2023}, the ideal real-world research dataset in clinical domain. We generated 10,000 synthetic discharge notes ($\mathcal{N}_S$). 

After we generated the discharge notes, we then generated the questionnaire ($Q$) and the conversation history (${Conv}_S$) between the educator and the patient using the prompt in Figure~\ref{fig:synthetic-qa-generation-prompt} and also the questionnaires using the prompt illustrated in Figure~\ref{fig:synthetic-conv-generation-prompt}. Here we also instruct GPT-4o-mini to generate the datasets. In our instructions, we include the discharge note ($\mathcal{N}^{i}_{S}$) and the medical conversation strategies that are listed in Table~\ref{tab1-medcontent-strategy}. All of the dataset were written in English.

\subsection{Evaluation for Synthetic Discharge Notes}
\begin{figure*}
    \centering
    \includegraphics[width=\linewidth, trim=0cm 0cm 14cm 0.8cm,clip]{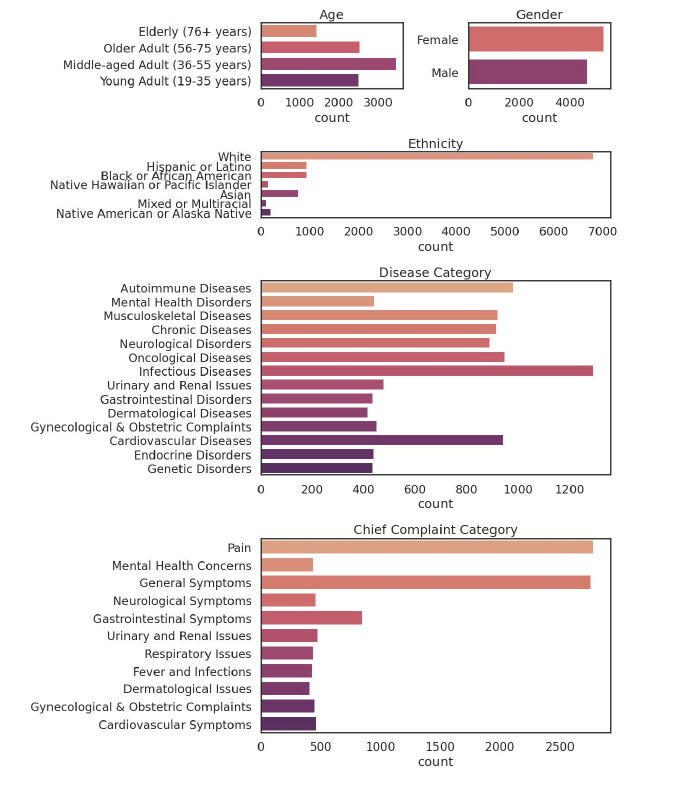}
    \caption{Demographic distribution of $\mathcal{N}_S$.}
    \label{fig:silver-demographic}
\end{figure*}

To ensure the quality of the dataset, we performed quality check measures for $\mathcal{N}_S$. We first analyzed the distribution of the dataset to verify the diversity (Figure~\ref{fig:silver-demographic}). As seen in the figure, we have successfully diversified the contents of the discharge note using strict guidelines when instructing GPT-4o-mini to generate synthetic discharge note. We also conducted a case analysis on the generated discharge note. As seen in figure~\ref{fig:silver-example-discharge-note}, the note contains the six medical contents that should appear in an ideal discharge note--Return to the Hospital/ED, medication, Diagnosis, Post-discharge treatment, Test and treatments during stay and Follow-up information-these are highlighted in the figure. By carefully coordinated demographic and clinical combinations, we ensure the generated discharge notes are clinically relevant and also diverse. 

\begin{figure*}
    \centering
\begin{tcolorbox}[width=\linewidth,
                  % enhanced,
                  % sharp corners,
                  %%frame hidden,
                  % interior hidden,
                  % halign=flush left,
                  boxsep=0pt,
                  left=5pt,
                  right=5pt,
                  top=5pt,
                  title=Guidelines for Annotation,
                  ]%%
% \centering

1. You are going to create 5-10 questions for each discharge note. \\
2. These questions are going to be clinically “relevant” and also important for the patient. \\
3. What is concerned “relevant” is as follows : \\
 i) It has to be acknowledged in the discharge note \\
 ii) It has to be concerned with the current health issues for that particular stays \\
 iii) It has to be concerned with instructions from the medical doctor \\
 iv) The categories that you could consider. The questions could be asked from in such categories :  \\
\quad Diagnosis during hospital stay \\
Procedure(interventions/tests) during hospital stay \\
Medication during hospital stay \\
Diagnosis in discharge \\
Procedure(follow up/tests/interventions) after discharge \\
Medication after discharge \\

Example questions : \\
Q. Why were you admitted to the hospital? \\
Q. What is the medication that the doctor recommended you to take? \\
Q. To treat your <illness/symptom> what drug did the doctor prescribe you? \\
Q. During your stay, the staff found you had <illness/symptom>. What was the name of that illness? \\
Q. The Doctor warns about your danger of <illness/symptom>. What kind of treatment/intervention did he recommend? \\
Q. What was your diagnosis during your stay? \\
Q. What is the cause of your symptoms? \\
Q. What is the correct dose of Gabapentin? \\
Q. What is the purpose of taking Benzonatate 100 mg three times a day as needed for cough? \\
Q. What procedure was performed during your hospital stay? \\
Q. What is the dosage of Lantus at night? \\

4. What is NOT considered “relevant” is as follows :  \\
 i) It does not appear in the discharge notes and cannot be inferred from the discharge notes \\
 ii) If it has less issues with the current health state of the patient or if it’s something that happened in the past that does not affect current health related concerns \\

5. How to comprise the choices
 i) you will come up with 3 choices for each questions
 ii) each choices will be either answer, distractor and irrelevant choice
 iii) distractor can be defined as something similar to the answer that causes confusion but not the actual answer that the question is looking for. E.g. distractors that are opposite to the answer would be one example.
 iv) irrelevant choice should be something that is bizarre, out of context. It should appear in the discharge note, but a totally irrelevant answer to the question.\\

\end{tcolorbox}
\caption{Guidelines for initial questionnaire generation for $Q$}
\label{fig:guideline-generation-for-Q}
\end{figure*}

\begin{figure*}
    \centering
\begin{tcolorbox}[width=\linewidth,
                  % enhanced,
                  % sharp corners,
                  %%frame hidden,
                  % interior hidden,
                  % halign=flush left,
                  boxsep=0pt,
                  left=5pt,
                  right=5pt,
                  top=5pt,
                  title=Guidelines for Annotation,
                  ]%%
% \centering

1. You are going to evaluate 5-10 questions for each discharge note. \\

2. These questions are going to be clinically “relevant” and also important for the patient. \\

3. What is concerned “relevant” is as follows : \\
 i) It has to be acknowledged in the discharge note \\
 ii) It has to be concerned with the current health issues for that particular stays \\
 iii) It has to be concerned with instructions from the medical doctor \\
 iv) The categories that you could consider. The questions could be asked from in such categories :  \\
Diagnosis during hospital stay \\
Procedure(interventions/tests) during hospital stay \\
Medication during hospital stay \\
Diagnosis in discharge \\
Procedure(follow up/tests/interventions) after discharge \\
Medication after discharge \\
 
4. How to \\
i) If you think the question is okay, please check relevant. \\
ii) If you consider that the question itself needs to be totally removed or changed please check irrelevant.  \\
iii) if you consider the question is okay but needs some modification please check modify and leave a comment below how we should change the questions \\
iv) if you checked irrelevant or modify please write what should be changed and guidance on how to fix the text or the question. \\

\end{tcolorbox}
\caption{Guidelines for questionnaire modification for $Q$}
\label{fig:silver-expert-annotation-guideline}
\end{figure*}

% $$\mathcal{D}_{Silver} = \{(\mathcal{N}^{i}_{S}, \mathcal{C}^{i}) | i \in \text{[1,10,000]} \}$$ 
% \begin{figure*}
% \onecolumn
\begin{figure*}
\begin{tcolorbox}[
    title=Synthetic note generation prompt,
    % enhanced jigsaw,
    % breakable, 
    % height fill
    ]

You are an expert in medicine with a lot of experience.
Please generate a synthetic Electronic Health Record (EHR) discharge notes for a scenario that a patient is discharging from a hospital. You will be given some basic demographic information. Please generate according to these predetermined information. \\

Demographic : 

  Disease category : \{disease category\}\\
  Age category : \{age\} \\
  Sex : \{sex\} \\
  Ethnicity : \{ethnicity\} \\
  Chief Complaint category : \{chief complaint\} \\
  Procedures : \{ procedure\} \\

The notes should contain the following subjects: \\
1) Indications to return to the Hospital/ED: Sign/Symptoms that the patient should be aware of when that person should contact or return to the hospital/Emergency Department. \\
2) Medication Information: The medication that the patient takes post-discharge.  \\
3) Diagnosis: The chief complaint of the patient, the main and sub diagnosis of the patient. This should be in Unified Medical Language System (UMLS) vocabulary. \\
4) Post-discharge treatments: What kind of actions or activities that the patient should be or should not be doing post-discharge.  \\
5) treatments/tests during stay: What type of treatment/tests were done during their stay, and what the results were.  \\
6) Follow up: When and where the patient should be following up the patient's health issues post-discharge.  \\

The format of the note should be as follows: 

Note ID : [note id]       

Sex: [sex] \hspace{3cm}              Chief Complain: [chief complaint of the patient] 

Past Medical History: [Past medical diagnosis] \\
Family History: [Family history] \\
Social History: [Social history] \\

1. Patient Summary \\
% [Brief Patient Summary]

2. Patient History \\
% [Brief Patient History]

3. Procedures and Progress during stay \\
% [Tests/Procedures and their results during stay]

4. Discharge Instructions  \\
% : [Discharge Diagnosis, Discharge Vital Signs, Discharge Disposition/Facility, Discharge Medications, Discharge instructions]

5. Discharge Summary \\
% : [discharge summary]

|||END 
...
% Make sure that the notes are medically consistent and correct. 
% You should be filling the parts with the brackets. e.g.[]
% Do not create any bold texts.
% Do not create any names or any mistakes that would violate privacy restrictions.

\end{tcolorbox}
\caption{Synthetic note generation prompt}
\label{fig:synthetic-note-generation-prompt}
\end{figure*}

\begin{figure*}
\begin{tcolorbox}[title=Synthetic questionnaire generation prompt]

You are an expert and an educator in medical domain.
You will be given a patient's discharge note. Your task is to generate 10 questionnaire for the discharge note which you think is important that the patient knows. It should be a multi-choice questionnaire where one is the answer, two of them are irrelevant, distractors. Please make sure that the question contents include the following topics : \\

Medical Contents: \\
1) Indications to return to the Hospital/ED: Sign/Symptoms that the patient should be aware of when that person should contact or return to the hospital/Emergency Department.\\
2) Medication Information: The medication that the patient takes post-discharge. \\
3) Diagnosis: The chief complaint of the patient, the main and sub diagnosis of the patient. This should be in Unified Medical Language System (UMLS) vocabulary. \\
4) Post-discharge treatments: What kind of actions or activities that the patient should be or should not be doing post-discharge. \\
5) treatments/tests during stay: What type of treatment/tests were done during their stay, and what the results were.  \\
6) Follow up: When and where the patient should be following up the patient's health issues post-discharge. \\

Here are some example questions. Note that you don't have to follow exactly what it says here, but this is just to give you a general idea what kind of questions you should make.  \\
Example Questions: \\
 What is your diagnosis? \\
 What treatments or procedures did you receive? \\
 What medications were prescribed, and what are they for? \\
 How should you take your medications, including dosage and timing? \\
 What are the possible side effects of your medications? \\
 ...

The format of the output should be in a list of jsons. \\

% [ \\
% \{\{"question" : "what is your discharge diagnosis?", \\
% "choices" : \{\{ \\
% 	"a" : "...", \\
% 	"b" : "...", \\
% 	"c" : "..."\}\}, \\
% "answer" : [choose between "a", "b", "c"], \\
% \}\}, \\
% \{\{"question" : "what is medication A for?", \\
% "choices" : \{\{ \\
% 	"a" : "...", \\
% 	"b" : "...", \\
% 	"c" : "..."\}\}, \\
% "answer" : [choose between "a", "b", "c"], \\
% \}\} \\
% ... \\
% ]
...

Please provide your response solely in the list of json format without including any text. Do not omit any braces. Do not include any text or code fences (like ```). The JSON must be valid and properly closed with \}\}.

Discharge note :
\{discharge\_note\}

Output :

\end{tcolorbox}
\caption{Synthetic questionnaire generation prompt}
\label{fig:synthetic-qa-generation-prompt}
\end{figure*}

\begin{figure*}
    
\begin{tcolorbox}[title=Synthetic Conversation history generation prompt]

You are an expert in medical domain. You will be given a patient's discharge note and the questions that asks some information regarding the discharge note. Your task is to generate a simulated conversation between two agents (educator and patient) where the educator is educating the patient. Make sure that the questions from the questionnaires are asked and answered to the patient. Patient may or may not know the answer to those questions.
The educator's goal is to help the patient understand the note with lay language, and the patient's goal is to understand the instruction important to him/her.  \\

Please make sure that the educated contents follow the conversation strategies provided below : \\
1) Fostering relationship: Build rapport and connection, Respect patient statements, privacy, autonomy, Engage in partnership buildiing. Express caring and commitment. Use appropriate language. Encourage patient participation. Show interest in the patient as a person. \\
2) Gathering information: Attempt to understand the patient's needs for the encounter. Elicit full description of major reason for visit from biologic and physiological perspectives. Ask open-ended questions. Allow patient to complete responses. Listen actively. Elicit patient's full set of concerns. Elicit patient's perspective on the problem/illness. Explore full effect of the illness. Clarify of the information. Inquire additional concerns. \\
3) Providing information: Seek to understand patient's informational needs. Share information. Overcome barriers to patient understanding. Facilitate understanding. Explain nature of the problem and approach to diagnosis, treatment. Give uncomplicated explanations and instructions. Avoid jargon and complexity. Encourage questions and check understanding. Emphasize key messages. \\
4) Decision making: Outline collaborative action plan. Identify and enlist resources and support. Discuss follow-up and plan for unexpected outcomes. \\
5) Enabling disease and treatment-related behavior: Assess patient's interest in and capacity for self-management. Provide advice (information needs, coping skills, strategies for success). Agree on next steps. Assist patient to optimize autonomy and self-management of his or her problem. Arrange for needed support. Advocate for, and assist patient with, health system. Assess patient's readiness to change health behaviors. Elicit goals, ideas, and decisions. \\
6) Responding to emotions: Facilitate patient expression of emotional consequences of illness. Acknowledge and explore emotions. Express empathy, sympathy, and reassurance. Provide help in dealing with emotions. Assess psychological distress. \\

Please also provide evidence from the original note for every physician chatbot's utterance. 
Follow the following format to construct your output. \\
...
% examples : \\
%  assistant: Hi, let's begin to go over important points in your discharge notes. \\
%  user : Okay, what should I know? \\
%  assistant : Do you know what your discharge diagnosis is? \\

% The format of the output should be in json format. \\
% ...
% Please provide your response solely in the json format without including any text. Do not omit any braces. Do not include any text or code fences (like ```). The JSON must be valid and properly closed with \}\}.\\

Discharge note : \\
\{discharge\_note\} \\

Questionnaire : \\
\{Questionnaire\} \\

Output : \\
\end{tcolorbox}
\caption{Synthetic conversation history generation prompt}
\label{fig:synthetic-conv-generation-prompt}
\end{figure*}

% \twocolumn
% \end{figure*}

% \begin{table}[]
%     \caption{Demographic Category }
%     \label{tab:silver-note-demographic}
%     \centering
%     \begin{tabular}{lll}
%     \toprule
%          Category & Contents & Ratio\\
%          \midrule
%          Age & Young Adult (19-35 years) & 0.250 \\
%           & Middle-aged Adult (36-55 years) & 0.350 \\
%           & Older Adult (56-75 years) & 0.250\\
%           & Elderly  (76+ years) & 0.150 \\
%          Gender & Male & 0.471 \\
%                 & Female &  0.529 \\
%          Ethnicity & White & 0.672 \\
%          & Black or African American & 0.100 \\
%          & Hispanic or Latino & 0.100 \\
%          & Asian & 0.080 \\
%          & Native American or Alaska Native & 0.020 \\
%          & Native Hawaiian or Pacific Islander & 0.015\\
%          & Mixed or Multicultural & 0.013\\
%     \bottomrule
%     \end{tabular}
% \end{table}

\begin{table*}[]
    \caption{Clinical combinations for generating $\mathcal{N}_S$}
    \begin{adjustbox}{width=\linewidth}
    \label{tab:silver-note-combinations}
    \centering
    \begin{tabular}{p{0.25\linewidth}p{0.3\linewidth}p{0.4\linewidth}}
    \toprule
         Disease Category & Chief Complaints & Associated Procedures \\
         \midrule
         Infectious Diseases & Fever and Infections, Respiratory Issues, Gastrointestinal Symptoms & Medication, Laboratory test, Vital Sign measurement \\
         Chronic Diseases & Pain, General symptoms & Medication, Physical therapy, Surgery, Diagnostic Imaging, Laboratory test, Vital Sign measurement  \\
         Cardiovascular Diseases & Cardiovascular symptoms, Pain & Cardiac Catheterization, Physical Therapy, Diagnostic Imaging, Laboratory test, Vital Sign measurement, Medication \\
         Neurological Disorders & Neurologic Symptoms, Pain & Physical Therapy, Diagnostic Imaging, Laboratory test, Vital Sign measurement, Medication \\
         Mental Health Disorders & Mental health concerns & Medication, Laboratory testing, Vital Sign measurement \\
         Oncological Diseases & Pain, General symptoms & Surgery, Chemotherapy, Radiation therapy, Medication, Laboratory testing, Vital Sign measurement \\
         Autoimmune Diseases & Pain, General symptoms & Medication, Laboratory testing, Vital Sign measurement \\
         Genetic Disorders & General symptoms & Medication, Laboratory testing, Vital Sign measurement \\
         Endocrine Disorders & General symptoms & Medication, Laboratory testing, Vital Sign measurement \\
         Musculuskeletal Disorders & Pain, General symptoms & Physical therapy, Surgery, Medication, Laboratory testing, Vital Sign measurement \\
         Gastrointestinal Disorders & Gastrointestinal symptoms & Endoscopy, Medication, Laboratory testing, Vital Sign measurement \\
         Dermatological Disorders & Dermatological issues & Wound care, Medication, Laboratory testing, Vital Sign measurement \\
         Urinary and Renal Disorders & Urinary and Renal issues & Dialysis , Medication, Laboratory testing, Vital Sign measurement \\
         Gynecological \& Obstetric issues & Gynecological \& Obstetric complaints & Surgery, Diagnostic Imaging, Medication, Laboratory testing, Vital Sign measurement \\
    \bottomrule
    \end{tabular}
    \end{adjustbox}
\end{table*}

% \onecolumn
% \twocolumn

% \onecolumn
\begin{figure*}
\sethlcolor{lime}
\centering
\begin{tcolorbox}[width=\linewidth,sharp corners,
    boxsep=0pt,
    left=5pt,
    right=5pt,
    top=5pt,
    % breakable,
]

Note ID:123456 \hspace{4em} Name:John Doe \\ 
Sex:Male\hspace{4em}          	Chief Complaint:Shortness of breath \\
Past Medical History:Hypertension, Type 2 Diabetes Mellitus  \\
Family History:Father with coronary artery disease, mother with hypertension  \\
Social History:Smokes 1 pack per day, occasional alcohol use, lives alone  \\

--- \\

1. Patient Summary \\
... \\

% The patient is a 65-year-old male admitted for acute exacerbation of chronic obstructive pulmonary disease (COPD) characterized by shortness of breath, increased sputum production, and wheezing. The patient has a history of smoking and poorly managed diabetes. \\

% 2. Patient History  \\
% The patient presented to the emergency department with complaints of progressive shortness of breath over the past week. The patient has a history of COPD and was previously stable but noted an increase in frequency of inhaler use in the past month. He reported adherence to prescribed medications but admitted to missed appointments for diabetes management. \\

\hl{3. Procedures and Progress during stay   \\
During the hospital stay, the following tests and treatments were performed:
- Spirometry on admission showed moderate obstruction, FEV1 55\% of predicted. \\
- Chest X-ray was unremarkable for acute processes. \\
- Blood tests revealed hyperglycemia (BG 250 mg/dL) and elevated White Blood Cell count. \\
- The patient received nebulizer treatments every 4 hours and was placed on systemic corticosteroids. \\
- Antibiotics were initiated due to a suspected respiratory infection, and upon clinical improvement, continued for a total of 7 days.
The patient’s respiratory status improved with treatment, and he was weaned off supplemental oxygen.} \\

4. Discharge Instructions \\

\hl{Discharge Diagnosis:Acute exacerbation of chronic obstructive pulmonary disease (COPD)} \\
...\\ 
% Discharge Disposition/Facility: Discharged to home with self-care instructions  \\

\hl{Discharge Medications:
- Albuterol 90 mcg, 2 puffs every 4-6 hours as needed for wheezing \\
- Prednisone 40 mg orally once daily for 5 days \\
- Metformin 500 mg orally twice daily for diabetes management \\
- Amlodipine 5 mg orally once daily for hypertension} \\

Discharge instructions: \\
- \hl{Indications to return to Hospital/ED: The patient should return to the hospital or contact a healthcare provider if he experiences increased shortness of breath, chest pain, persistent cough with blood, fever greater than 101°F, or signs of an allergic reaction to medications (e.g., rash, swelling).} \\
- \hl{Post-discharge treatments: The patient should continue all prescribed medications, avoid smoking and any exposure to respiratory irritants, and maintain a low-sugar diet with ample hydration. Engage in light physical activity as tolerated but avoid strenuous activities until cleared by a follow-up physician.} \\
- \hl{Follow-up: The patient is advised to schedule a follow-up appointment with his primary care physician within 1 week for management of COPD and diabetes.} \\

5. Discharge Summary \\
The patient was admitted for acute exacerbation of COPD and has shown significant improvement after received treatment. Upon discharge, he was educated about medication adherence and lifestyle modifications necessary for better management of his respiratory condition and diabetes. He was informed about signs and symptoms that would warrant further medical attention.  \\

|||END
\end{tcolorbox}
\caption{Example of synthetic discharge note from $\mathcal{N}_S$}
\label{fig:silver-example-discharge-note}
\end{figure*}

% \twocolumn

\section{Experimental details}
\label{appendix:experiment-details}

\subsection{Data splits}
For ${Comp}_{S}$, we split into 8000, 1900, 100 ratio. Here, 8000 samples of $\mathcal{N}_S$ and ${Comp}_S$ were used in the Supervised Fine-Tuning. 1900 samples were used in validation to know whether the model is being overfitted or underfitted in the training process. The last 100 samples were used as the test set. For ${Comp}_{G}$, the notes were also used as a test set. 

\subsection{Reward Function}
\label{appendix:reward-function}

\begin{equation}
R = \frac{1}{T}\sum^T_{t=1}r_t
\label{eq:reward_function}
\end{equation}

Here, $r_t$ is the score for question $t$, assigned as 1 if the question is answered correctly and 0 otherwise. $T$ denotes the total number of questions in the test. For each medical note, we generated 5 to 10 multiple-choice questions, each with three possible answer options. During the PPO stage, the model receives a reward based on the number of questions it answers correctly—these points are accumulated and used as the reward signal for training.

\subsection{Generation configurations}
Our overall pipeline used huggingface transformer package and the vllm for implementation. We set the maximum sequence length for LLaMA 3.2-3B model to 60,000 token length. Also for each utterance generation we set 200 tokens as a maximum number of tokens to be generated. The temperatures were set to 0.6 for closed source models, and 0.2 for open-source models.  

\subsection{Evaluation prompts}

% \onecolumn
\begin{figure*}
    \centering
\begin{tcolorbox}[title=Medical content evaluation prompt, width=\linewidth]
\#\#\# Instruction :
You are an expert in clinical conversation. 
Here are the types of conversation categories between a physician and the patient. 
Please classify each given sentence based on these criteria. Below is the criterias and their explanation. Each sentence could have multiple categories of contents in it. 
We also give you some examples.  \\

\#\#\# Define conversation categories \\
1. Return to the ED/Hospital indications (c1) : If the conversations is about sign or symptoms when the patient should return to the ED/Hospital, then it should be classified as c1 \\
2. Medication Info (c2) : If the conversation is about a drug information that the patient is taking, then classify as c2. \\
3. Diagnosis (c3) : If it's about the diagnosis/disease of the current or past disease classify this as c3.  \\
4. Postdischarge treatment (c4) : If the conversation is about a treatment that is taken after the patient is discharged from the ED/Hospital then classify as c4. \\
5. tests and treatments (c5) : If the conversation is about a test/treatment that happened during the patient's stay then classify as c5.  \\
6. Follow-up (c6) : If the conversation is about where or when they will follow-up with their health issues then classify as c6.  \\
7. No matching (NA) : If there are no matching for the above criteria, classify as NA. \\

\#\#\# Example: \\

\#\#\# Sentence  \\
Hi How are you today? \\

\#\#\# Classifcation : NA \\

\#\#\# Sentence  \\
Got it. What about the bowel symptoms they mentioned? What should I look out for? \\

\#\#\# Classifcation : c3 \\

... \\

\#\#\# Sentence : \\
\{sentence-needs-to-be-examined\} \\

\#\#\# Classification : \\

Please output the class and no other strings included \\
\end{tcolorbox}
    \caption{Medical Content evaluation prompt}
    \label{fig:medical-content-evaluation-prompt}
\end{figure*}
% \twocolumn

% \onecolumn
\begin{figure*}
    \centering
\begin{tcolorbox}[title=Medical conversation strategy evaluation, width=\linewidth]
You are a medical expert who wants to evaluate how helpful and clinically appropriate a conversation between an agent and a patient is.  
You will be evaluating the conversation strategy specifically. 
Here is the conversation that we use to evaluate.
The patient asks some questions regarding their discharge notes and the agent answers, in order to help patients understand and memorize their discharge instructions.  \\

Six evaluation aspects for the agent's conversation strategy. \\

\textbf{Fostering relationship}: Build rapport and connection, Respect patient statements, privacy, autonomy, Engage in partnership building. Express caring and commitment. Use appropriate language. Encourage patient participation. Show interest in the patient as a person. \\
\textbf{Gathering information}: Attempt to understand the patient's needs for the encounter. Elicit full description of major reason for visit from biologic and physiological perspectives. Ask open-ended questions. Allow patient to complete responses. Listen actively. Elicit patient's full set of concerns. Elicit patient's perspective on the problem/illness. Explore full effect of the illness. Clarify of the information. Inquire additional concerns. \\
\textbf{Providing information}: Seek to understand patient's informational needs. Share information. Overcome barriers to patient understanding. Facilitate understanding. Explain nature of the problem and approach to diagnosis, treatment. Give uncomplicated explanations and instructions. Avoid jargon and complexity. Encourage questions and check understanding. Emphasize key messages. \\
\textbf{Decision making}: Outline collaborative action plan. Identify and enlist resources and support. Discuss follow-up and plan for unexpected outcomes. \\
\textbf{Enabling disease and treatment-related behavior}: Assess patient's interest in and capacity for self-management. Provide advice (information needs, coping skills, strategies for success). Agree on next steps. Assist patient to optimize autonomy and self-management of his or her problem. Arrange for needed support. Advocate for, and assist patient with, health system. Assess patient's readiness to change health behaviors. Elicit goals, ideas, and decisions. \\
\textbf{Responding to emotions}: Facilitate patient expression of emotional consequences of illness. Acknowledge and explore emotions. Express empathy, sympathy, and reassurance. Provide help in dealing with emotions. Assess psychological distress. \\

5-point likert scale:  \\
1: very low rating \\
2: low rating \\
3: neutral or medium rating \\
4: higher rating \\
5: very highly rating \\

The conversation between the patient and the AI model:  \\
\{conversation-history\} \\

Give the 5-point likert scale of the agent's conversation quality (six aspects) one by one. When providing the evidence, please describe what would help to improve the score to make them the full 5 point. Keep the evidence concise and short. \\
...
% Return the scores as json format, adhering to the following structure: 

% \{\{
%  "Fostering relationship": \{\{"score": ..., "evidence": ...\}\},
%  "Gathering information": \{\{"score": ..., "evidence": ...\}\},
%  ...
% \}\}

% Please provide your response solely in the json format without including any text. Do not omit any braces. Do not include any text or code fences (like ```). The JSON must be valid and properly closed with \}\}. \\
\end{tcolorbox}
    \caption{Medical conversation strategy evaluation prompt}
    \label{fig:medical-conversation-strategy-evaluation-prompt}
\end{figure*}

% \twocolumn

The evaluation prompts used for medical content evaluation and medical conversation strategy evaluation are in Table~\ref{fig:medical-content-evaluation-prompt} and Table~\ref{fig:medical-conversation-strategy-evaluation-prompt}, respectively. 

\subsection{Hardware settings \& Compute time}

The server has Intel(R) Xeon Gold 6226R CPU @ 2.90GHz. It required 4 Nvidia Quadro RTX 6000 24GB to fine-tune LLaMA 3.2-3B-Instruct. Fine-tuning took around 2 hours using the training set, for the reinforcement learning stage, around 1-2 hours. However, it took more than suggested compute time for the actual experiments since we had unsuccessful experiments to find the appropriate hyperparameter for the generation configurations.  

\section{Calculating Confidence Intervals}
For each test set of $\mathcal{N}_G$ and $\mathcal{N}_S$, there are 100 independent samples. And for each of these instances, we measured the simulated conversation's score. Each value is reported as mean ± margin of error, where the margin is computed as $t_{\alpha/2} \cdot \text{SEM}$ using 100 samples and a 95\% confidence level. We report the margin error for Table~\ref{tab:content-gold-silver} and Table~\ref{tab:strategy-gold-silver}. For Table~\ref{tab-generation-metric-results}, we report a single shot tryout result.

\section{Evaluation for generation metrics}
We used Huggingface's BLEU, ROUGE-L, BERTscore and NLTK library's Flesch-Kincaid Grade Level api to calculate the generation metrics.

\section{Medical content evaluation}
\label{appendix:medical-content-evaluation}

As illustrated in Table~\ref{tab:content-case}, the evaluation results of using LLM-as-a-judge can easily classify the relevant medical content categories of the utterance of the models. This is not a difficult task that requires difficult reasoning such as mathematics. The outputs are highly reliable.

\definecolor{LightCyan}{rgb}{0.88,1,1}
\begin{table}[ht]
    \caption{Case study with LLM-as-a-judge for medical content.}
    \label{tab:content-case}
    \begin{adjustbox}{width=\columnwidth, center}    
    \centering
    \begin{tabular}{p{0.04\linewidth}p{0.85\linewidth}}
        \toprule
        \multicolumn{2}{l}{Medical content evaluation outputs} \\
        \midrule
        % 1 & \textbf{Chatbot} : Absolutely! Following up after three months means we'll check again to make certain everything heals properly inside your belly. Does that sound right? \\
        % \rowcolor{LightCyan}
          % & \textbf{LLM-as-a-judge} : Follow-up \\
        % \midrule
        1 & \textbf{Chatbot} : Great question! One serious symptom would require calling our doctors immediately—namely worsening persistent pain despite proper management. Got that covered? \\
        \rowcolor{LightCyan}
          & \textbf{LLM-as-a-judge} : Return to Hospital/ED \\
        \midrule
        2 & \textbf{Chatbot} : One thing definitely worth keeping track of isn’t feeling fatigued anymore quickly enough within those first few weeks. That’s really good news! Do remember that? \\
        \rowcolor{LightCyan}
          & \textbf{LLM-as-a-judge} : Diagnosis \\
        \midrule
        3 & \textbf{Chatbot} : Yeah absolutely--it helps prevent complications related specifically due infections developing near-the-incised-area-so-staying-cautious-is-vital-right-now!? \\
        \rowcolor{LightCyan}
          & \textbf{LLM-as-a-judge} : Post-discharge treatment \\
        \midrule
        4 & \textbf{Chatbot} : Don't Take More Medicine Than Prescribed Even When Felt Severe Because Over-Dosing Has Serious Consequences Always Better Safe Then Sorry!!. \\
        \rowcolor{LightCyan}
          & \textbf{LLM-as-a-judge} : Medication \\
        \midrule
        5 & \textbf{Chatbot} : During our next visit together, one important thing would be getting another chest x-ray before then; does that sound right to you? \\
        \rowcolor{LightCyan}
          & \textbf{LLM-as-a-judge} : Test and Treatments, Follow-up \\
        \bottomrule
    \end{tabular}
    \end{adjustbox}
\end{table}

\section{Medical conversation strategy evaluation}
\label{appendix:medical-conversation-strategy}

The case study results are shown in Table~\ref{tab:case-medical-strategy}. This is a fully automated scoring system, therefore we compared the output of the model and the analysis of an expert. In general, we found that the evaluation has minor discrepancies between the LLM-as-a-judge and the human expert evaluation results. We found some differences in Enabling disease and treatment-related behavior and responding to emotions. But the overall trend has a similarity. This way, we justify the quality of the LLM-as-a-judge based evaluation for medical conversation strategy.

\begin{table*}[h]
  \caption{Case study with LLM-as-a-judge for medical conversation strategy for a conversation between our model and the patient agent}
  \label{tab:case-medical-strategy}
  \centering
  \resizebox{\linewidth}{!}{
  \begin{tabular}{p{0.35\linewidth}p{0.05\linewidth}p{0.05\linewidth}p{0.5\linewidth}}
    \toprule
         % \multicolumn{3}{l}{Medical conversation strategy evaluation output} \\
         % \midrule
         Category & LLM Score & Expert Score & Evidence \\
         \midrule
        Fostering relationship & 4/5 & 4/5 & The agent exhibited caring and engaged with the patient, but further personalization and acknowledgment of the patient's feelings would strengthen the rapport.  \\
        Gathering information & 4/5 & 4/5 & The agent asked appropriate questions and listened actively; however, encouraging more open-ended responses would deepen understanding of the patient's concerns. \\
        Providing information & 4/5 & 5/5 & The agent provided clear and understandable instructions, but occasional jargon and complex phrasing detracted from clarity. \\
        Decision making & 3/5 & 3/5 & The agent discussed follow-up and assured the patient but could better outline collaborative decision-making and resource identification. \\
        Enabling disease and treatment-related behavior & 4/5 & 3/5 &  
        The agent provided useful advice but could further enhance the patient's autonomy by discussing self-management strategies more explicitly. \\
        Responding to emotions & 3/5 & 5/5 &  
        While the agent acknowledged some emotions, more empathetic engagement and probing into the patient's feelings could improve emotional support. \\
        \bottomrule
    \end{tabular}
    }
\end{table*}

% \twocolumn

\section{Turing test details}
\label{appendix:human-evaluation}

% \subsection{Details for the human evaluation}
% \begin{figure}
%     \centering
%     \includegraphics[width=0.8\linewidth, trim=1cm 4cm 1cm 0cm, clip]{human-experiment-framework.pdf}
%     \caption{Human evaluation}
%     \label{fig:human-evaluation}
% \end{figure}

\subsection{Information for the participants}

\begin{figure*}[h]
    \centering
\begin{tcolorbox}
This study aims to test the robustness of healthcare AI agents in discharge scenarios. You will be either taking the role of educator or the patient. We randomly assigned each of you to a role and will let you know which role you are assigned to. \\

For the educators you will be asked to stay in room A, and for the patient roles, you will be asked to stay in room B. For nurse educators, you will be asked to stay at room C.  \\

For the educator role, your goal is to deliver as much information that is written in the discharge note as possible. Prioritize the things that should be taught first, such as discharge diagnosis, medication information, discharge instructions and such. For the patient role, you will be asked to engage with your educator, learn and remember as much information as possible. This will last around 15 minutes.  \\

After the engagement, educator roles can leave the room and finish their participation. The patient will take a comprehension test that has 8-10 questions about your discharge note. This will last for 15 minutes.  \\

After the test is finished, we will collect your test sheet and let you know whether you were in group A,B or C. And whether you engaged with a real human being or a chatbot.  \\
\end{tcolorbox}
    \caption{Information given to human subjects}
    \label{fig:information-for-participants}
\end{figure*}

Instructions given to the human subjects can be seen in figure~\ref{fig:information-for-participants}. Each participants were randomly assigned to their roles. For the patient roles, the identity of their counterpart, the educator, was not disclosed until the study was finished. 

\subsection{Enrollment and Experiment}

\begin{figure*}
    \centering
    \includegraphics[width=\linewidth, trim=1cm 0cm 2cm 1cm,clip]{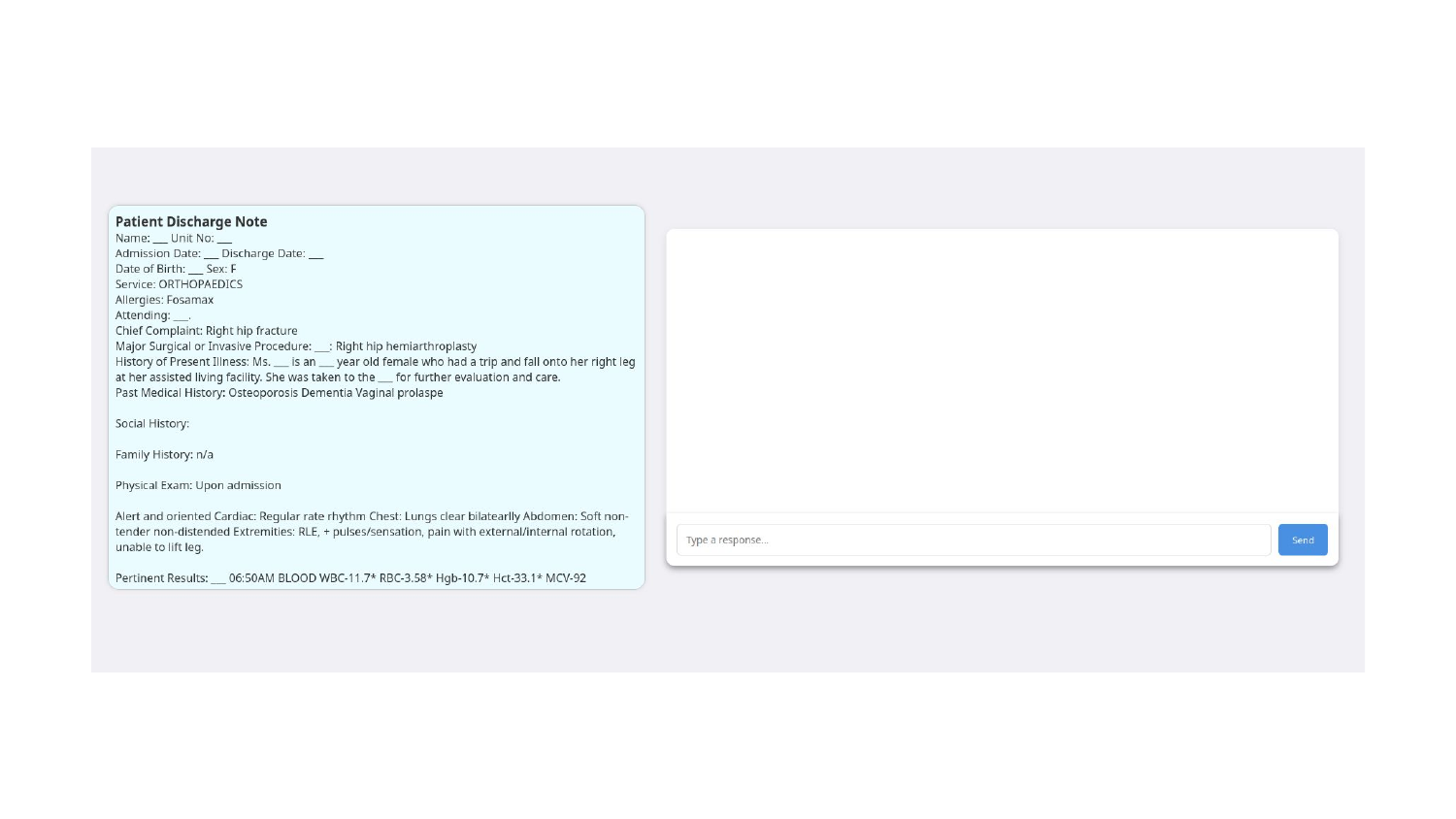}
    \caption{Interface for experts for the Turing test}
    \label{fig:interface-expert}
\end{figure*}

\begin{figure*}
    \centering
    \includegraphics[width=\linewidth, trim=2cm 0cm 3cm 1cm, clip]{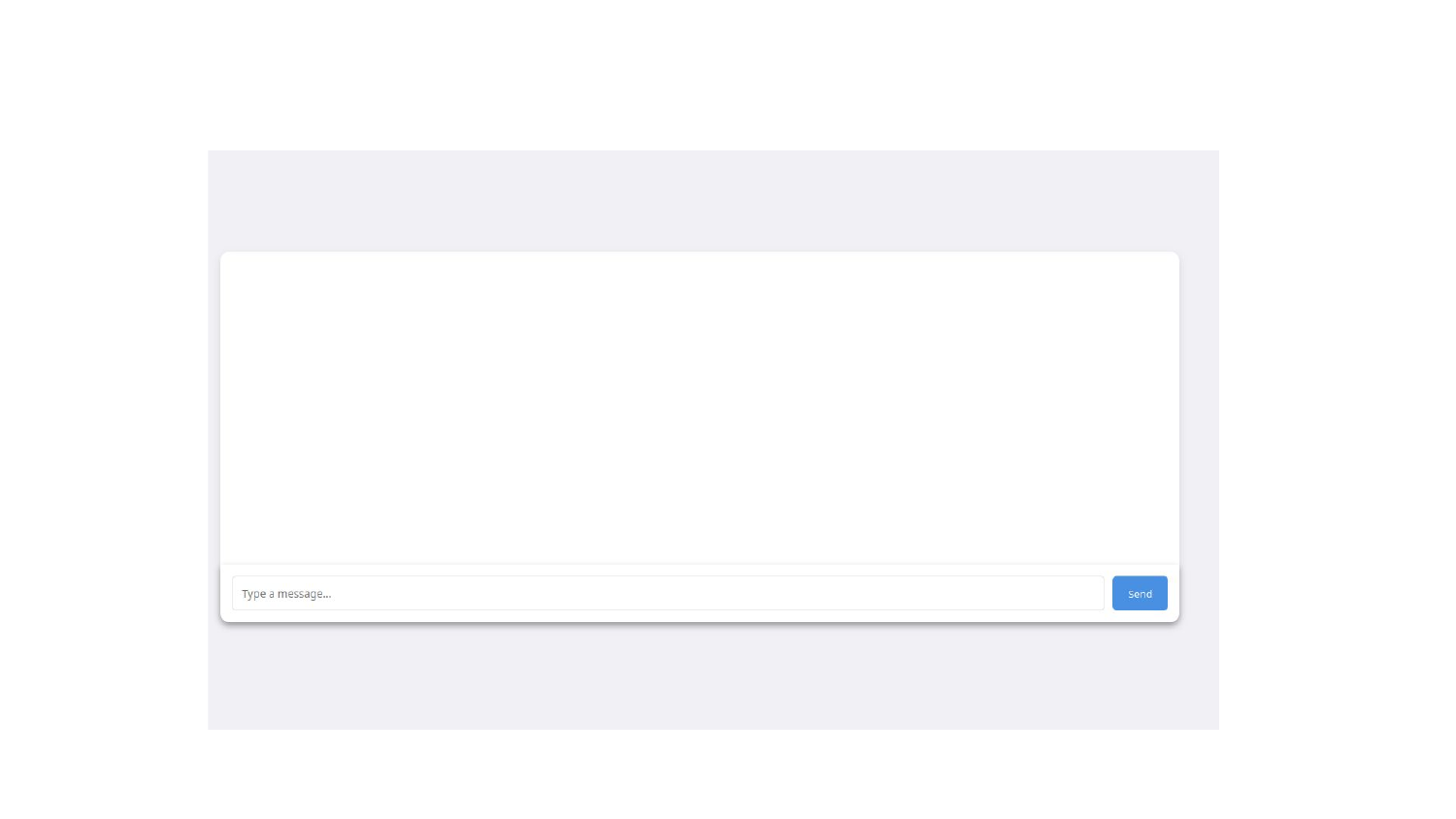}
    \caption{Interface for patients for the Turing test}
    \label{fig:interface-patient}
\end{figure*}
In our enrollment process, we first explained the experiment and then took their verbal consent. After 2 weeks, we randomly assigned the role for each participants. Asking them to prepare their pen, and laptop. Using their laptop, depending on their role, they were asked to login as an expert or the patient as seen if Figure~\ref{fig:interface-expert}, and Figure~\ref{fig:interface-patient}. The participants taking the patient role, did not know whether their educator was going to be a human or not. They interacted for 15 minutes regarding the discharge note.

After the dialogue took place, the patient roles were asked to take the comprehension test and guess whether their counterpart felt like human or not.

\subsection{Data Confidentiality and Potential Privacy Risk}

In order to protect the personal information of the participants we removed the collected participant's personal information and only used their subject number in processing the results. We eliminated their names and emails to protect the participant's privacy.

\section{Failure Case Analysis}

\begin{figure}[h]
    \centering
    \includegraphics[width=\columnwidth, trim=0cm 0cm 0cm 0cm, clip]{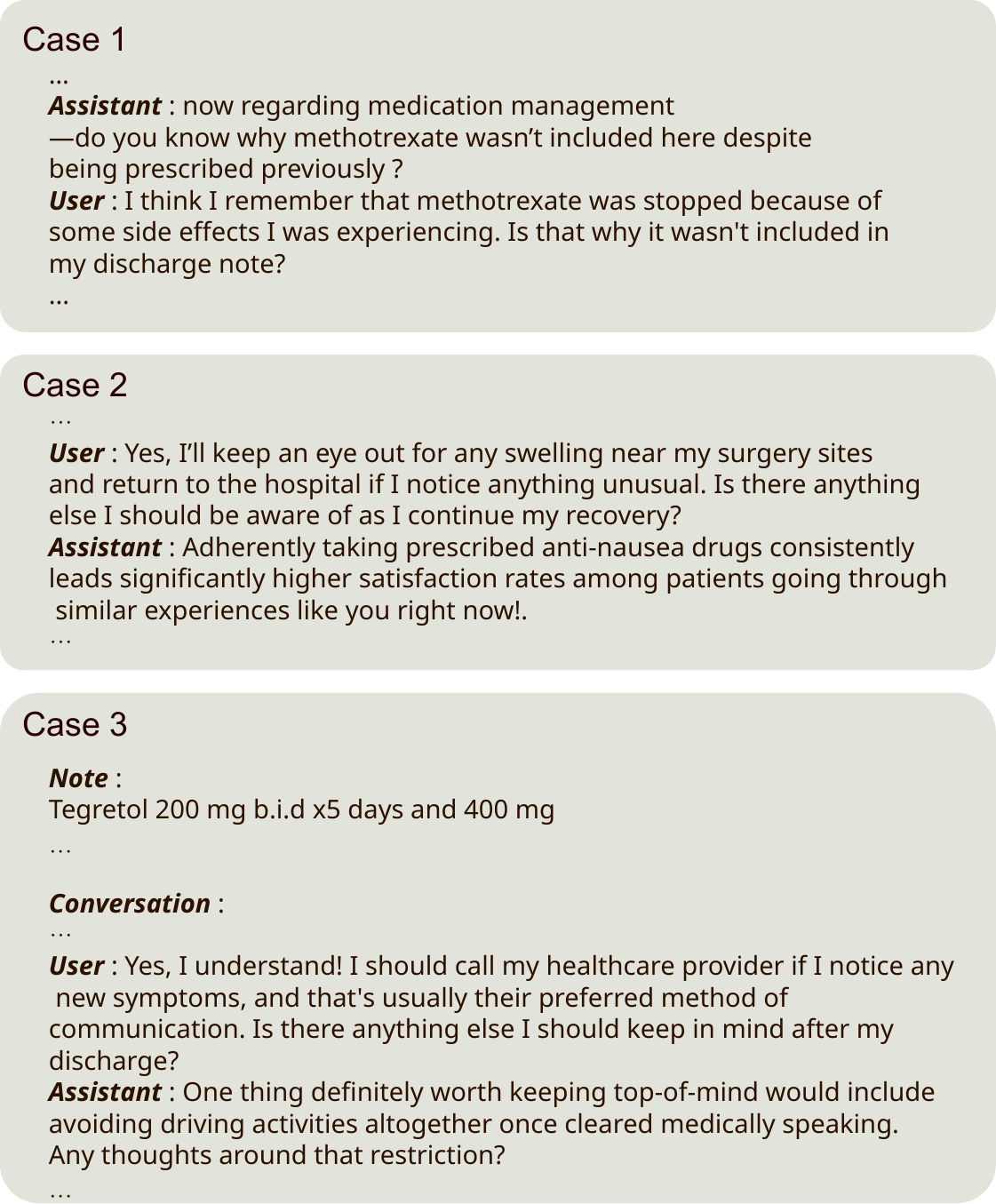}
    \caption{Three examples of failed cases of our trained model. Our model similar to other LLMs suffers hallucination issues.}
    \label{fig:failure-case}
\end{figure}

We also share the cases where the model fails to acknowledge the shortcomings of the framework~\ref{fig:failure-case}. In the first case (Case 1), the model asks why it didn’t have methotrexate prescription. However, in the medical note, there was no mention of methotrexate. Although there was a similar drug but different drug named “Methadone”. We have found several cases in the dialogue between the chatbot and the patient agent of showing these behaviors. We noticed these behaviors occurs when there are similar drug names, or the format of the discharge note is different from the discharge note that it was already trained on. These are the ones considered as "dangerous" hallucinations which could have devastating effects to the patient. In the second case (Case 2), the model recommends anti-nausea medications, despite the fact that none were prescribed in the discharge note. While the patient is presumed to have cancer and be undergoing treatment—conditions where anti-nausea medication may be clinically appropriate for managing treatment-related side effects—this information was not explicitly stated in the discharge summary. In contrast to the first case, this example presents a scenario of a "positive" or minimally harmful hallucination, prompting the question of whether strict adherence to the discharge note always represents the optimal approach. In the third case (Case 3), although not mentioned in the discharge note, the model by its own knowledge recommends some of the precautions to the patient. The only medication listed is Tegretol, which is related to the patient’s recent stroke. While there is no explicit instruction to avoid driving while taking this medication, the model advises the patient to refrain from such activities. Although this may be a reasonable precaution in this context, offering unsolicited advice without clinical backing can lead to unintended consequences in other cases.

Overall, this underscores the critical importance of addressing the risks involved in deploying language models in patient-facing applications without appropriate safeguards. Case studies such as this illustrate the need for implementing robust guardrails and precautionary measures to protect patient safety. Designing mechanisms to prevent potential harms should be a central focus in the development of patient education systems. Our future research will also focus on developing tools and criteria aimed at enhancing patient safety.

% \twocolumn

% \section{Broader Impact}
% \label{appendix:broader-impact}

% Our study has several societal implications. Our study shows that utilizing carefully curated dataset with reinforcement learning based alignment can result in lightweight, domain-specific models that are robust. In some domains, there are problems of data scarcity, but these issues can be mitigate using the similar methods in our study. However, since the process also relies on the robustness and the quality of the synthetically generated dataset, expert domain are required to overlook the process for quality control measures.   

\end{document}